\documentclass[10pt,journal,compsoc]{IEEEtran}

\usepackage{graphicx}
\usepackage{makecell}
\usepackage{tikz}
\usepackage{comment}
\usepackage{amsmath,amssymb} 
\usepackage{color}
\usepackage{booktabs}
\usepackage{arydshln}
\usepackage{multirow}
\usepackage{hyperref}

\usepackage{tikz}

\usepackage{graphicx}
\usepackage[caption=false,font=normalsize,labelfont=sf,textfont=sf]{subfig}

\renewcommand{\paragraph}[1]{\noindent 
 \textbf{#1}~}

\newcommand{\blue}[1]{\textbf{\textcolor{mblue}{#1}}}
\newcommand{\red}[1]{\textcolor{red}{#1}}
\newcommand{\bred}[1]{\textbf{\textcolor{red}{#1}}}
\newcommand{\green}[1]{\textcolor{mgreen}{#1}}
\newcommand{\yellow}[1]{\textcolor{yellow}{#1}}
\newcommand{\gray}[1]{\textcolor{mgray}{#1}}

\newcommand{\textbfg}[1]{\textbf{\textcolor{mgreen}{#1}}}

\newcommand{\frag}[0]{\textbf{{\textit{fragments}}}}

\definecolor{mgray}{gray}{0.35}
\definecolor{mred}{RGB}{238, 34, 12}
\definecolor{mgreen}{RGB}{1, 127, 0}
\definecolor{mblue}{RGB}{0, 0, 180}

\ifCLASSOPTIONcompsoc
  \usepackage[nocompress]{cite}
\else
  \usepackage{cite}
\fi

\begin{document}
%
\title{Neighbourhood Representative Sampling for Efficient End-to-end Video Quality Assessment}

\author{Haoning Wu,
        Chaofeng Chen,
        Liang Liao,~\IEEEmembership{Member,~IEEE},
        Jingwen Hou,~\IEEEmembership{Student Member,~IEEE}, \\
        Wenxiu Sun,
        Qiong Yan,
        Jinwei Gu,~\IEEEmembership{Senior Member,~IEEE},
        Weisi Lin,~\IEEEmembership{Fellow,~IEEE}
\IEEEcompsocitemizethanks{\IEEEcompsocthanksitem H. Wu, C. Chen, L. Liao are with S-Lab, Nanyang Technological University (NTU), Singapore (email: haoning001@e.ntu.edu.sg, [chaofeng.chen, liang.liao]@ntu.edu.sg). \protect
\IEEEcompsocthanksitem J. Hou and W. Lin are with School of Computer Science and Engineering, Nanyang Technological University (NTU), Singapore (jingwen003@e.ntu.edu.sg, wslin@ntu.edu.sg).\protect
\IEEEcompsocthanksitem W. Sun, Q. Yan and J. Gu are with Tetras. AI and Sensetime Research ([sunwx, yanqiong]@tetras.ai, gujinwei@sensebrain.ai).\protect
\IEEEcompsocthanksitem Corresponding author: Weisi Lin.}
\thanks{Preprint Edition. Under Review.}}

\markboth{Under Review for IEEE Transactions on Pattern Analysis and Machine Intelligence}%
{Shell \MakeLowercase{\textit{et al.}}: Bare Advanced Demo of IEEEtran.cls for IEEE Computer Society Journals}

\IEEEtitleabstractindextext{%
\begin{abstract}
The increased resolution of real-world videos presents a dilemma between efficiency and accuracy for deep Video Quality Assessment (VQA). On the one hand, keeping the original resolution will lead to unacceptable computational costs. On the other hand, existing practices, such as resizing and cropping, will change the quality of original videos due to the loss of details and contents, and are therefore harmful to quality assessment. 
With the obtained insight from the study of spatial-temporal redundancy in the human visual system and visual coding theory, we observe that quality information around a neighbourhood is typically similar, motivating us to investigate an effective quality-sensitive neighbourhood representatives scheme for VQA. In this work, we propose a unified scheme, spatial-temporal grid mini-cube sampling (St-GMS) to get a novel type of sample, named \frag. Full-resolution videos are first divided into mini-cubes with preset spatial-temporal grids, then the temporal-aligned quality representatives are sampled to compose the fragments that serve as inputs for VQA. In addition, we design the Fragment Attention Network (FANet), a network architecture tailored specifically for fragments. With fragments and FANet, the proposed efficient end-to-end \textbf{FAST-VQA} and \textbf{FasterVQA} achieve significantly better performance than existing approaches on all VQA benchmarks while requiring only \textbf{1/1612} FLOPs compared to the current state-of-the-art. Codes, models and demos are available at \url{https://github.com/timothyhtimothy/FAST-VQA-and-FasterVQA}.

\end{abstract}

}

\maketitle

\IEEEdisplaynontitleabstractindextext

%
\IEEEpeerreviewmaketitle

\ifCLASSOPTIONcompsoc
\IEEEraisesectionheading{\section{Introduction}\label{sec:introduction}}
\else
\section{Introduction}
\label{sec:introduction}
\fi


\maketitle

\IEEEPARstart{V}{isual} content with a large spatial resolution has always been the pursuit of humans.
Indeed, with the proliferation of high-definition photographing devices and significant advancements in various technologies such as video compression and 4G/5G, the videos shot by most common users have greatly increased in resolution (\textit{e.g.}, 1080P, 4K, or even 8K), thereby largely enriching human perception and entertainment styles.  Nevertheless, the increased size of real-world videos has posed a number of practical obstacles for machine algorithms in terms of capture, transmission, storage, analysis, and evaluation of those videos. Video Quality Assessment (VQA), also known as the quantification of human perception of video quality, severely suffers from the growing video sizes.

\begin{figure}[t]
    \centering
    \includegraphics[width=\linewidth]{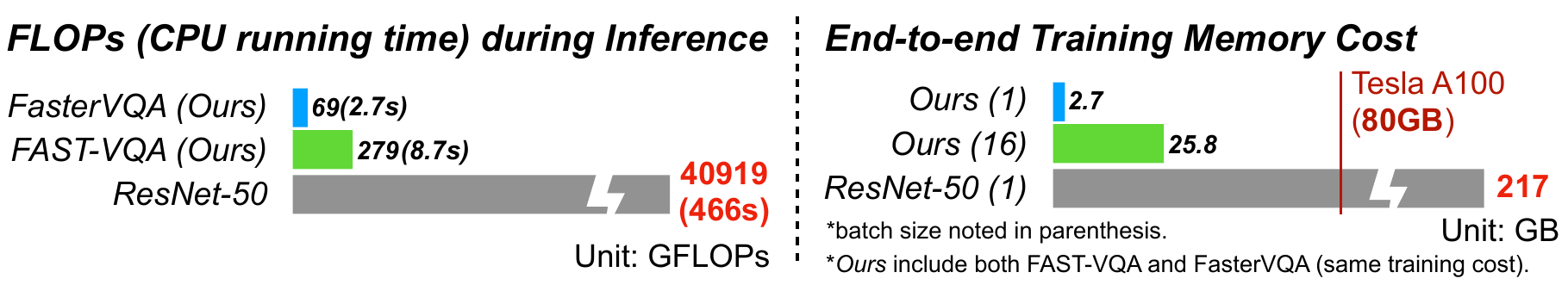}
    \vspace{-17pt}
    \caption{Inference cost (FLOPs, running time) and training memory cost of a vanilla ResNet-50 on a full 1080P, 10-second-long video (without any sampling), compared with our methods (FAST-VQA/FasterVQA).}
    \label{fig:0}
    \vspace{-15pt}
\end{figure}

 While classical shallow VQA algorithms \cite{vbliinds,viideo,tlvqm,videval} based on handcrafted features struggle to handle in-the-wild videos with diverse contents and degradation types, the most recent and effective approaches on in-the-wild VQA are based on deep neural networks \cite{vsfa,mdtvsfa,pvq,mlsp,cnn+lstm,cnntlvqm}. However, the computational complexity of deep neural networks usually grows with the video size, \textit{i.e.,} quadratically with the resolution, making them intolerable on high-resolution videos. Taking a 10-second-long 1080P video clip as an example, a plain ResNet-50\cite{he2016residual} as the network backbone will require \textbf{40,919GFLOPs} computational cost for inference and \textbf{217GB} graphic memory cost during training with a batch size of 1 (Fig.~\ref{fig:0}), which exceeds the memory limits of all GPUs at present. In order to alleviate computational resource and memory shortage issues on GPUs, the majority of deep VQA methods \cite{vsfa,lsctphiq,mlsp,mdtvsfa,pvq,bvqa2021,discovqa} choose to regress quality scores with \textit{{{fixed}} features} extracted from pre-trained networks of classification tasks~\cite{he2016residual,irnv2,r3d} instead of end-to-end training, resulting in these methods lacking effective representation learning and essentially only training a shallow regressor for VQA.

\begin{figure*}[htbp]
    \centering
    \includegraphics[width=0.96\linewidth]{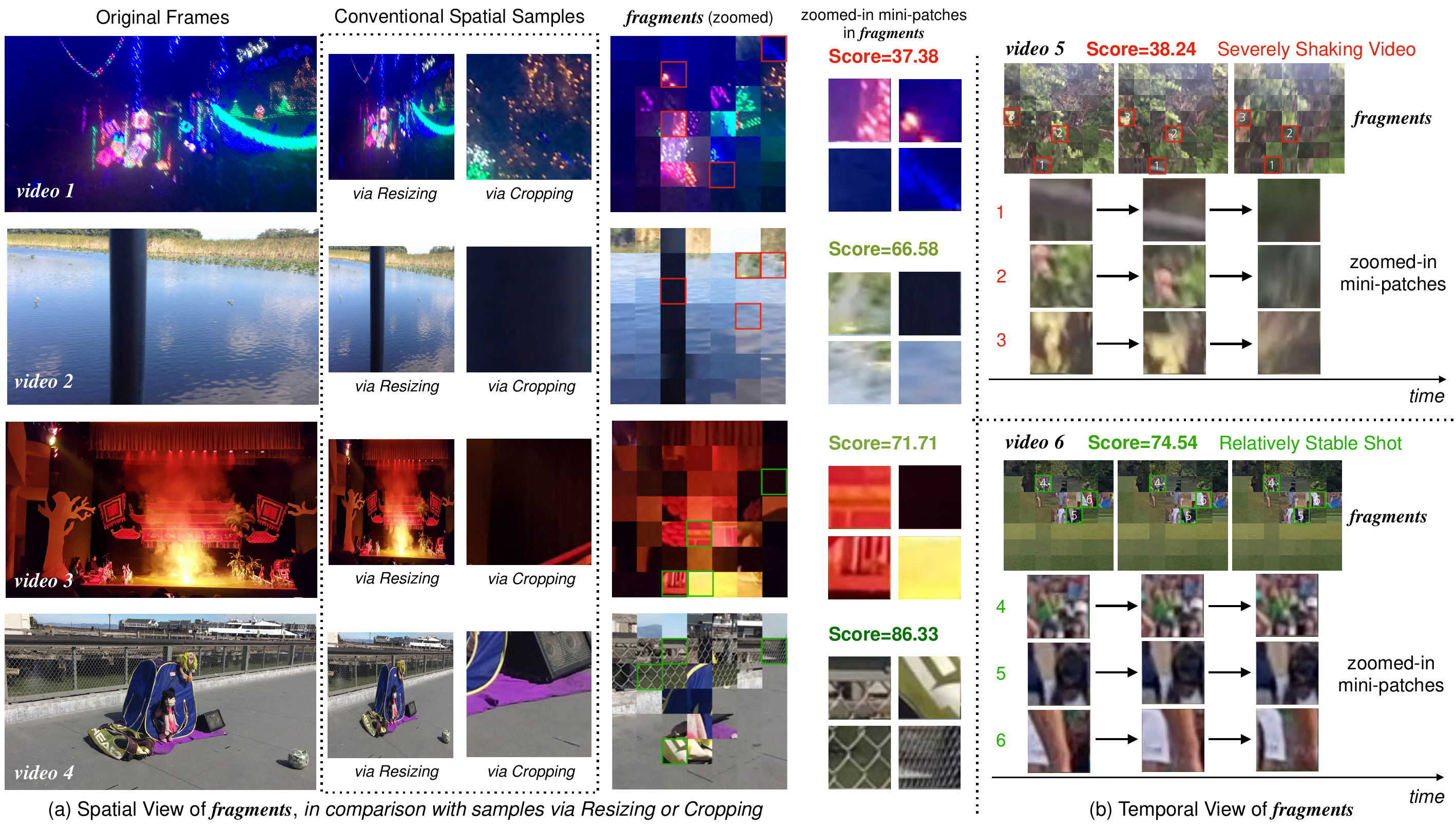}

    \vspace{-11pt}
    \caption{\textbf{\textit{Fragments}}, in spatial view (compared with resizing and cropping) (a) and temporal view (b). Zoom-in views of mini-patches show that \textbf{\textit{fragments}} can retain spatial local quality information (a), and spot temporal variations such as shaking across frames (b).}
    \label{fig:2}
    \vspace{-12pt}
\end{figure*}

Meanwhile, some other video-related tasks employ various sampling strategies to avoid the high computational cost. Most of them obtained their insight from studies on the human visual system (HVS) \cite{2015humanbrain} or visual coding theories \cite{epsr1,epsr2,eptr1}, which proved that visual content tends to be similar around a local region, \textit{i.e.,} a neighbourhood. For example, image and video compression standards, \textit{e.g.,} JPEG~\cite{jpeg} and H264/AVC \cite{h264}, and resizing algorithms, such as Bicubic \cite{bicubic}, generally extract representatives for partitioned neighbourhoods to ensure that the resampled information can represent the original information. As a result, most high-level video recognition (\textit{e.g.,} classification, detection) methods \cite{swin3d,k400,mvit,vivit} have adopted resizing to reduce the video dimensions. However, as illustrated in Fig.~\ref{fig:2}(a), resizing corrupts quality-related local textures such as blurs and artifacts in \textit{video 1}\&\textit{2} which is significant in VQA and other low-level tasks. On the other hand, in order to preserve these local textures, several works \cite{crop1,crop2} attempt to crop a single continuous patch. Nevertheless, these samples lose a large proportion of quality information, \textit{e.g.,} \textit{video 2}\&\textit{3} in Fig.~\ref{fig:2}(a), thus also not suitable for the VQA task. To build good samples for VQA, we need to ensure that they are representative of global quality information while also preserving the sensitivity to quality information on local textures and temporal variations.

{In this paper, we propose a new sampling paradigm to tackle with VQA, \emph{quality-sensitive neighbourhood representatives}, that only requires sampling representatives from partitioned neighbourhoods but also selects texture-sensitive raw continuous patches as representatives}. Specifically, we design a unified spatial-temporal sampling scheme, Spatial-temporal Grid Mini-cubes Sampling (St-GMS). Spatially, it cuts video frames into uniform non-overlapping grids, and samples a mini-patch randomly from each grid. Temporally, it cuts videos into uniform segments and samples multiple continuous frames within each segment. To better preserve temporal continuity between frames, we also constrain that mini-patches in each spatial grid and temporal segment should be aligned to form a mini-cube. Finally, all the mini-cubes are stitched to an integrated sample specially designed for VQA, termed \frag~(Fig.~\ref{fig:2}).

Fig.~\ref{fig:2}(a) illustrates the spatial view of \textbf{\textit{fragments}}. First, they preserve the local texture-related quality information (\emph{e.g.}, spot blurs happened in \textit{video 1}\&\textit{2}) by retaining the patches in original resolution. Second, benefiting from the globally uniformly partitioned grids, {fragments} cover the global quality even though different regions have different qualities (\emph{e.g.}, \textit{video 2\&3}). Third, by splicing the mini-cubes, {{fragments}} retain contextual relations among them so that the model can learn global scene information and rough semantic information of the original frames. As for the temporal view of \textbf{\textit{fragments}}, as shown in Fig.~\ref{fig:2}(b), with the continuous frames and aligned mini-patches in each segment, {{fragments}} can also spot temporal variations in videos, \textit{e.g.}, distinguish between severely shaking videos (\emph{e.g.}, \textit{video 5}) from relatively stable shots (\emph{e.g.}, \textit{video 6}). The segment-wise sampling on the temporal dimension also ensures temporally uniform coverage of quality information.

It is non-trivial to design deep networks for \textbf{\textit{fragments}}, as the mini-cubes are actually independent and the edges in between may be misinterpreted as quality defects. To avoid uncontrolled fusion of pixels in different mini-cubes, we propose a rule for building networks on \textbf{\textit{fragments}}, the \textit{match constraint}, to align the pooling operations with sampled mini-cubes. Specifically, we choose Video Swin Transformer \cite{swin3d} as the backbone and improve the Relative Position Biases in the backbone into Gated Relative Position Biases (GRPB) to correctly represent the positions of pixels in \textit{\textbf{fragments}}. Based on the characteristic of \textbf{\textit{fragments}} that quality is diverse among mini-cubes, we further replace the pool-first head that is usually used in high-level tasks with a pool-last Intra-Patch Non-linear Regression (IP-NLR) head, to get better performance and predict local quality maps beyond quality scores. In general, with a Tiny Swin Transformer (\textit{abbr.} as Swin-T) as baseline backbone and the proposed GRPB \& IP-NLR modules as modifications, we propose the Fragment Attention Network (FANet) that best extracts the quality-sensitive information in \textbf{\textit{fragments}}.

This work is a substantial extension to our earlier conference version FAST-VQA\cite{fastvqa} which proposes a spatial-only sampling scheme and the accommodated network structure (FANet). In comparison to the conference version, we include a significant amount of improvements:
\textbf{1)} To further improve efficiency, we extend spatial-only sampling in to the spatial-temporal sampling scheme (St-GMS), based on which we improve FAST-VQA into the \underline{F}r\underline{a}gment \underline{s}pa\underline{t}ial-t\underline{e}mpo\underline{r}al \underline{V}ideo \underline{Q}uality \underline{A}ssessment (\textbf{FasterVQA}) that performs comparable to FAST-VQA with only 25\% of FLOPs
\textbf{2)} We propose the Adaptive Multi-scale Inference (AMI) on FANet for adaptively inferring on different scales with one model trained on a fixed scale while keeping competitive performance.
\textbf{3)} We add extensive ablation studies to further analyze the effects of sampling granularity, end-to-end training and semantic pre-training in the proposed methods.
The main contributions of this work are listed as follows:
\begin{itemize}

\item We propose the \textit{quality-sensitive neighbourhood representatives}, a novel sampling paradigm for VQA, and design a unified Spatial-temporal Grid Mini-cube Sampling (St-GMS) scheme to sample \frag. The fragments enable deep VQA methods to efficiently and effectively evaluate videos of any resolution.

\item We propose and evaluate the \textit{match constraint} for pooling layers as guidance for building networks for \frag. Based on this constraint, we propose the Fragment Attention Network (FANet) with newly designed GRPB and IP-NLR modules to best accommodate the characteristics of \frag.

\item The proposed FAST-VQA and FasterVQA outperform existing VQA methods by a large margin (up to 7\%) with unprecedented efficiency (up to $1612\times$). Our efficient version can even infer at 13.6$\times$ faster than real-time on CPU with competitive accuracy.


\end{itemize}

\section{Related Works}

\textbf{{Classical VQA Methods.}} Classical VQA methods \cite{niqe,bofqa,rrstedqa,diivine} employ handcraft features to evaluate video quality. Some methods hypothesize \cite{brisque,viideo,vbliinds,tpqi} that natural videos follow specific statistical rules, while the defect videos do not, and compute quality scores only from statistical evidence without regression from any subjective labels. In recent years, several methods \cite{tlvqm,stgreed,videval} choose to first handcraft quality-sensitive features and then regress them to subjective mean opinion scores (MOS), in order to better fit the human perception. Among them, TLVQM \cite{tlvqm} uses a combination of two levels of handcraft features, including high-complexity spatial features computed on sparse frames for measuring spatial distortions, and low-complexity temporal features computed for each frame for assessing temporal variations. VIDEVAL \cite{videval} ensembles various handcraft features to model the diverse authentic distortions and also reduces the feature dimensions to reduce the computational burden. Spatial-temporal chips are sampled in a recent work called ChipQA\cite{chipqa} for more efficient handcraft feature extraction. These classical approaches suggest that it is possible to reduce the size of videos while retaining their quality information. Nevertheless, since the factors affecting the in-the-wild video quality are quite complicated and usually cannot be concluded by finite handcraft features, the performance of these classical methods are constrained. 


\noindent \textbf{Deep VQA Methods.} Benefiting from the semantic awareness of deep neural network features, deep VQA methods \cite{cnn+lstm,deepvqa} are becoming predominant. For example, VSFA \cite{vsfa} uses the features extracted by pre-trained ResNet-50 \cite{he2016residual} from ImageNet-1k dataset \cite {imagenet} and adopts Gate Recurrent Unit (GRU)~\cite{gru} for quality regression. However, due to the extremely high memory cost of deep networks on high-resolution videos (as shown in Fig.~\ref{fig:0}), most existing deep VQA methods \cite{gstvqa, vsfa, mlsp, rirnet, dstsvqa} can only extract fixed features instead of updating them. Without end-to-end training, existing methods generally improve features in the three following ways. 1) Introducing heavier backbones, \textit{e.g.}, MLSP-FF~\cite{mlsp} includes heavier Inception-ResNet-V2 \cite {irnv2} for feature extraction. 2) Using multiple backbone networks instead of one, \textit{e.g.}, PVQ\cite{pvq} uses an additional ResNet-3d-18\cite{r3d} network to extract temporal quality features. 3) Including frame-wise pre-training \cite{lsctphiq,pvq,cnntlvqm} from IQA databases \cite{koniq,paq2piq}. A most recent method, BVQA-TCSVT-2022\cite{bvqa2021}, combines all these three ways to reach better performance, while it requires up to 26 minutes on CPU to assess the quality for an 8-second-long video, 200$\times$ slower than video playback. While improving performance, these practices significantly sacrifice the final computational efficiency. These practices further highlight the value of the proposed method with effective end-to-end training via efficient quality-retained sampling, so as to improve performance in an efficient manner for training and inference. 

\begin{figure*}[htbp]
    \centering
    \includegraphics[width=0.92\linewidth]{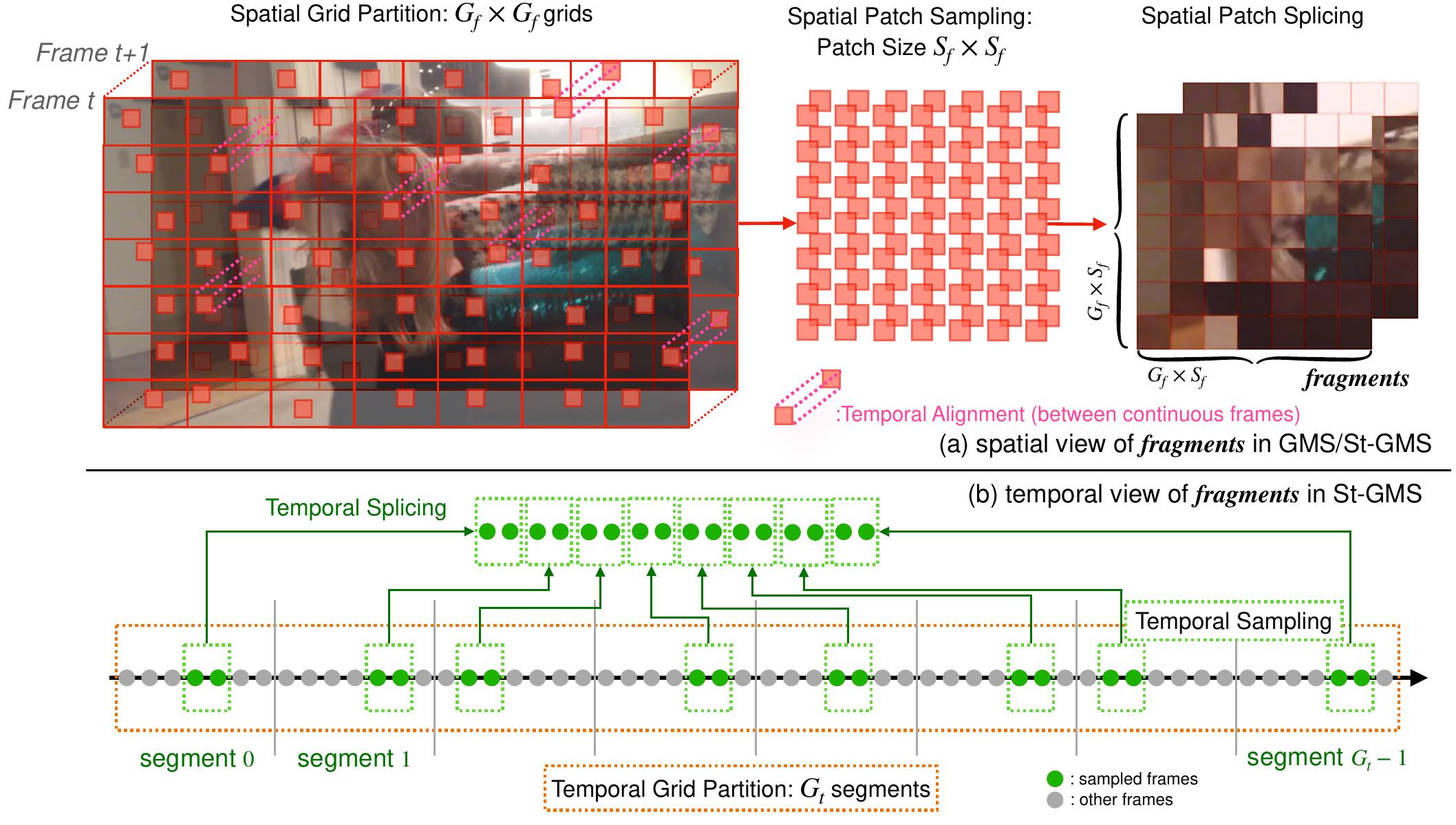}
    \vspace{-6pt}
    \caption{The pipeline for sampling \frag~with Spatial-temporal Grid Mini-Cube Sampling (St-GMS, Sec.~\ref{section:gms}), including spatial ({a}, discussed in Sec.~\ref{sec:gmss}) and temporal ({b}, discussed in Sec.~\ref{section:stgms}) sampling operations. The sampled \frag~are fed into the FANet (Fig.~\ref{fig:5}).}
    \label{fig:4}
    \vspace{-14pt}
\end{figure*}

\section{Approach}
\label{section:method}

In this section, we introduce the proposed FAST-VQA and FasterVQA. We first define the paradigm of sampling quality-sensitive neighbourhood representatives (Sec.~\ref{sec:nqr}), and introduce the corresponding Spatial-temporal Grid Mini-cube Sampling (St-GMS, Sec.~\ref{section:gms}) scheme to resample the videos into \textbf{\textit{fragments}}. After sampling, the \textbf{\textit{fragments}} are fed into the Fragment Attention Network (FANet, discussed in Sec.~\ref{section:network}) which is designed based on the \textit{match constraint}. We also propose an Adaptive Multi-scale Inference (AMI, Sec.~\ref{section:AMI}) strategy for adaptive-scale inference on the model trained at a single scale. Lastly, we present the associated objective functions (Sec.~\ref{Loss}) for model training.


\subsection{Sampling Representatives from Neighbourhoods}
\label{sec:nqr}


In visual tasks, sampling is widely applied. Specifically, uniform sampling schemes, such as spatial nearest/bicubic downsampling and temporal uniform sampling, are widely applied in high-level recognition tasks. In general, these methods can be concluded by two steps: 1) segmenting the image/video into various local areas (referred to as \textit{neighbourhoods}), and 2) sampling a representative from each neighbourhood. We conclude the overall unified paradigm as \textbf{neighbourhood representatives} ($\mathcal{R}$) which can be specified to either spatial or temporal dimensions. Given a target sampled size $S$ and a single representative size $S_r$, the paradigm first divides the visual contents into neighbourhoods $\mathcal{N}=\{n^i|i=0,1,2,\dots,\frac{S}{S_r}-1\}$, and then the neighbourhood representatives $\mathcal{R}$ can be formulated as,
\begin{equation}
    \mathcal{R}=\{r(n^i)|i=0,1,2,\dots,\frac{S}{S_r}-1\} 
    \label{eq:nr}
\end{equation}
where $r(n^i)$ denotes the function that samples a representative from neighbourhood $n^i$.

As {neighbourhood redundancy} also occurs for quality-related information, the neighbourhood representatives can also be applied to quality tasks. Nevertheless, according to many widely acknowledged studies \cite{videval,tlvqm,chipqa}, continuous local textures and local temporal variations are significant while evaluating video quality, which will be corrupted if we apply resizing or uniform frame sampling ($S_r=1$). With deep thinking of the requirements of VQA task, we propose to sample \textbf{quality-sensitive neighbourhood representatives} ($\mathcal{R}_q$), which should satisfy: 1) they should contain raw pixels in videos instead of pooled or averaged results; and 2) the raw pixels in one representative $r(n^i)$ should form a continuous patch or clip that is large enough to distinguish spatial or temporal local quality information. As a result, these representatives $\mathcal{R}_q$ can represent both the unbiased global quality information and the sensitive local quality information (\textit{e.g.}, spatial local textures, temporal variations among adjacent frames) that are vital for VQA.

\subsection{Spatial-temporal Grid Mini-cube Sampling}\label{section:gms}


We propose the uniform \textbf{Spatial-temporal Grid Mini-cube Sampling (St-GMS)} scheme which follows the principle of quality-sensitive neighbourhood representatives in both spatial and temporal dimensions. The pipeline for St-GMS is illustrated in Fig.~\ref{fig:4} and discussed as follows.

\subsubsection{Spatial sampling: Grid Mini-patch Sampling (GMS)}
\label{sec:gmss}

In the first part, we discuss the Grid Mini-patch Sampling (GMS, Fig.~\ref{fig:4}(a)), \textit{i.e.}, the spatial sampling operations in St-GMS, together with the corresponding principles.

\paragraph{Representing global quality: uniform grid partition.} To include each region for quality assessment and uniformly assess quality in different areas, we design the grid partition to cut each video frame into uniform grids with each grid having the same size (as shown in Fig~\ref{fig:4}(a)). In particular, we cut the video frame $\mathcal{I}$ with size $H\times W$ into $G_f\times G_f$ uniform grids with the same sizes, denoted as $\{g^{i,j}\vert 0<i<G_f, 0<j<G_f\}$, where $g^{i,j}$ refers to the grid 
in $i$-th row and $j$-th column.  The partition is formalized as follows.\footnote{In this section, all square brackets ($_{[~]}$) denote the slicing operations, and all superscripts (\textit{e.g. $^i$}) denote position indices.}
{\begin{equation}
    g^{i,j} = \mathcal{I}_{[\frac{i\times H}{G_f}:\frac{(i+1)\times H}{G_f},\frac{j\times W}{G_f}:\frac{(j+1)\times W}{G_f}]}\label{eq:1}
\end{equation}}

\paragraph{Sensitive to local quality: raw patch sampling.} To preserve the local textures (\textit{e.g.}, blurs, noises, artefacts) that are vital in VQA, we select raw resolution patches without any resizing operations to represent local textural quality in grids. To keep sensitivity to local textures, we employ uniform random patch sampling to select one mini-patch $\mathcal{MP}^{i,j}$ of the size of $S_f \times S_f$ from each grid $g^{i,j}$. The spatial patch sampling ($\mathbf{S}_s$) is formulated as follows.

\begin{equation}
     \mathcal{MP}^{i,j} = \mathbf{S}_s^{i,j}(g^{i,j}),~~~~0\leq i, j < G_f\label{eq:2}
\end{equation}

\begin{figure*}[]
    \centering
    \includegraphics[width=\linewidth]{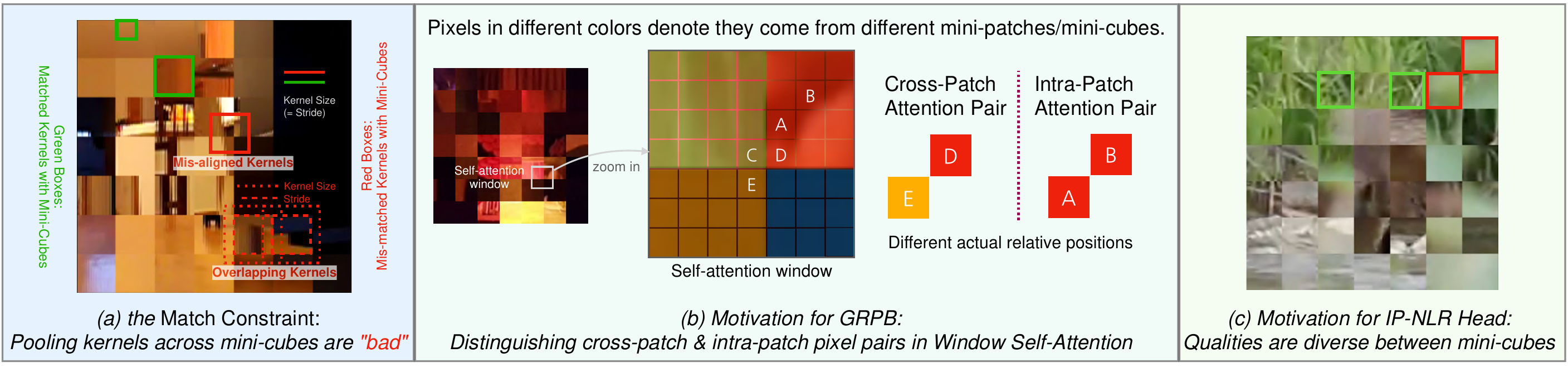}
    \vspace{-15pt}
    \caption{Motivation for the \textit{match constraint} (a) and two proposed modules in FANet: (b) Gated Relative Position Biases (GRPB); (c) Intra-Patch Non-Linear Regression (IP-NLR) head. The structure for the whole FANet is illustrated in Fig.~\ref{fig:5}.}
    \label{fig:3}
    \vspace{-12pt}
\end{figure*}

\paragraph{Preserving contextual relations: patch splicing.} Existing works~\cite{vsfa,mlsp,dbcnn} have shown that global scene information notably affects quality-related perception, that even the same textures under different semantic background can relate to different quality \cite{sfa}. To preserve the background information about the global scene, we retain the contextual relations among mini-patches by splicing them together:

\begin{equation}
\begin{aligned}
            \mathcal{F}^{i,j}&= \mathcal{F}_{ [i\times S_f:(i+1)\times S_f, j\times S_f:(j+1)\times S_f]} \\&= \mathcal{MP}^{i,j},\qquad\qquad 0\leq i, j < G_f \label{eq:3} 
\end{aligned}
\end{equation}
where $\mathcal{F}$ denotes the spliced mini-patches from frame $\mathcal{I}$ after spatial GMS pipeline, as in our conference version~\cite{fastvqa}. We also extend GMS into the temporal dimension for more efficient quality evaluation, discussed as follows.

\subsubsection{Extending GMS into the temporal dimension}\label{section:stgms}

We extend the GMS into the temporal dimension based on  \textbf{unified} quality-sensitive neighbourhood representatives, as illustrated in Fig.~\ref{fig:4}(b). We discuss the detailed principles and operations in the temporal dimension as follows.


\paragraph{Temporal representative: uniform segment partition.} Similar to the spatial case, an accurate VQA method also need to uniformly assess quality along the temporal dimension. For uniformity, TSN\cite{tsn} proposed general segment-wise sampling for videos which had been applied by many existing VQA methods \cite{videval,tlvqm,cnntlvqm}. Thus, we divide the video $\mathcal{V}$ with $T$ total frames into $G_t$ uniform non-overlapping temporal segments (as shown in Fig.~\ref{fig:4}(b)). Overall, we extend the uniform grid partition as defined in Eq.~\ref{eq:1} into spatial-temporal uniform grid partition, as follows.
\begin{equation}
g^{k,i,j} = \mathcal{V}_{[\frac{k\times T}{G_t}:\frac{(k+1)\times T}{G_t},\frac{i\times H}{G_f}:\frac{(i+1)\times H}{G_f},\frac{j\times W}{G_f}:\frac{(j+1)\times W}{G_f}]}
\end{equation}
where $g^{k,i,j}$ denotes the spatial-temporal grid in $k$-th temporal segment, $i$-th row and $j$-th column.

\paragraph{Sensitive to inter-frame variations: continuous frames.} It is widely recognized by early works \cite{deepvqa,tlvqm,pvq} that inter-frame temporal variations are influential to video quality. To retain the raw temporal variations in videos, we would like the frames sampled in each segment to be \textbf{continuous} and the corresponding mini-patches to be \textbf{aligned} so that the temporal variation inside the segment can be reflected by these samples.
Thus, we apply temporal continuous frame sampling ($\mathbf{S}_t$) before the raw-patch sampling ($\mathbf{S}_s$, Eq.~\ref{eq:2}) to sample a continuous \textbf{mini-cube} $\mathcal{MC}^{k,i,j}$ of size $T_f\times S_f\times S_f$ from each spatial-temporal grid $g^{k,i,j}$ as follows:
\begin{equation}
    \mathcal{MC}^{k,i,j}  = \mathbf{S}_{s}^{i,j} (\mathbf{S}^k_t(g^{k,i,j})),~~~~0\leq i, j < G_s, 0\leq k< G_t
\end{equation} 

\paragraph{Long-term dependencies: temporal splicing.}
Although there are no consensus on explanations of the long-term temporal dependencies in VQA, plenty of existing methods \cite{vsfa,lsctphiq,discovqa} have proved that they are practically influential to the video quality. Therefore, we include temporal splicing into the whole splicing operation as follows:
\begin{equation}
\begin{aligned}
\mathcal{F}_{\mathrm{3D}}^{k,i,j}
&=\mathcal{F} _{\mathrm{3D}[k\times T_f:(k+1)\times T_f, i\times S_f:(i+1)\times S_f, j\times S_f:(j+1)\times S_f]} \\
&= \mathcal{MC}^{k,i,j} ~~~~~~~ 0\leq i, j < G_s, 0\leq k < G_t
\end{aligned}
\end{equation}
where $\mathcal{F}_{3D}$ denotes the spliced spatial-temporal mini-cubes after the St-GMS pipeline, as space-time-unified \textbf{\textit{fragments}}. 

\begin{figure*}[]
    \centering
    \includegraphics[width=0.98\linewidth]{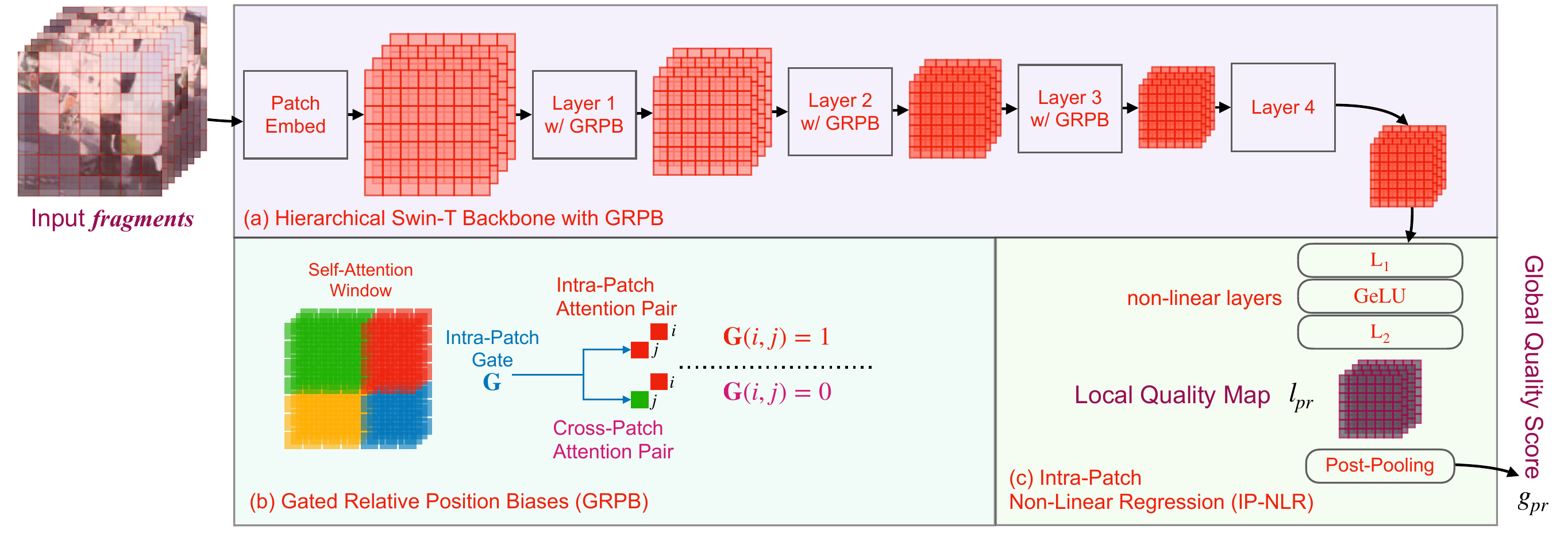}
    \vspace{-10pt}
    \caption{The overall framework for FANet, including the Gated Relative Position Biases (GRPB) and Intra-Patch Non-Linear Regression (IP-NLR) modules. The \frag~come from Grid Mini-patch Sampling (for FAST-VQA) or Spatial-temporal 
 Grid Mini-cube Sampling (for FasterVQA).}
    \label{fig:5}
    \vspace{-12pt}
\end{figure*}

The GMS and the following FANet (Sec.~\ref{section:network}, Fig.~\ref{fig:5}) together constitute the proposed FAST-VQA, which only includes the proposed spatial sampling operations and selects dense frames in the temporal dimension for inference. With unified spatial and temporal sampling strategies, we improve FAST-VQA into \textbf{FasterVQA} by replacing the GMS with the St-GMS. FasterVQA has \red{\textit{4X efficiency}} than FAST-VQA yet comparable accuracy. Both FAST-VQA and FasterVQA include the FANet structure, discussed as follows.

\subsection{Quality Regression Network for \textit{fragments}}\label{section:network}

\subsubsection{Motivation: \emph{match constraint} for pooling layers}

It is non-trivial to build a network using the proposed \frag~as inputs. Like most quality assessment networks, it should be able to effectively extract the quality information preserved in fragments, including the local textures inside mini-cubes and the contextual relationships between them. Moreover, it should specifically avoid misinterpreting the discontinuity between mini-cubes (resulted by artificial splicing) for local textures, which calls for more careful network design, especially for the \textbf{pooling layers} which decide the values of subsequent feature pixels and are not learnable. As a result, we impose the \textbf{\emph{match constraint}}, which constrains that each pooling kernel should only include pixels inside of an individual mini-cube as green boxes in Fig.~\ref{fig:3}(a)), but not between parts of mini-cubes (red boxes), before each mini-cube is finally downsampled as a single pixel. Formally, take any pooling kernel at any layer (before mini-cubes have been downsampled as single pixels), denote the set of original pixels that falls into the area of the kernel as $\mathcal{P}$, the constraint can be formulated as:
\begin{equation}
\label{eq:match}
\exists~~k,i,j,~~~~s.t.~~\mathcal{P} \subset \mathcal{MC}^{k,i,j}
\end{equation}

To follow the \emph{match constraint}, we require the networks that use non-overlapping pooling kernels. Many backbone structures can meet this requirement, including transformer-based structures \cite{swin3d,vivit,mvit,vit} and part of modern convolution-based structures such as ConvNeXt\cite{convnext}, while it is possible to match their pooling kernels with mini-cubes. Our experiments show that either \textbf{1)} using conventional backbones (\textit{i.e.,} ResNet~\cite{he2016residual} and MobileNet~\cite{mobilenetv3}) with overlapping pooling kernels or \textbf{2)} failing to align mini-cubes with pooled pixels leads to a notable performance drop, suggesting the significance of \emph{match constraint} for pooling layers. Finally, we choose the Video Swin Transformer Tiny (Swin-T) backbone which follows the \textit{match constraint} as the backbone of the quality regression network for \frag. We also make several modification on the Swin-T to better accommodate it for fragments, discussed as follows.

\subsubsection{Fragment Attention Network (FANet)}
\label{section:fanet}

\paragraph{The Overall Framework.} Fig.~\ref{fig:5} shows the overall framework of \textbf{Fragment Attention Network (FANet)}, the proposed end-to-end quality regression network for \frag. It includes a four-layer Swin-T with first three window self-attention layers modified by GRPB as the backbone (\textit{abbr. as} Swin-GRPB), and an IP-NLR quality-regression head.

\paragraph{Gated Relative Position Biases (GRPB).} In Swin-T, the window self-attention layers are built across mini-cubes to learn contextual relations between them. However, in these window self-attention layers, representing the positions of pixels of \textbf{\textit{fragments}} differs from those of normal inputs. While original Swin-T proposes relative position bias (RPB) that uses learnable Relative Bias Table ($\mathbf{T}$) to represent the relative positions of pixels in attention pairs ($QK^T$), they cannot well represent the relative positions of different pixels in \textbf{\textit{fragments}}. Specifically, considering that some pairs in the same attention window might have the same relative position (\textit{e.g.}, Fig.~\ref{fig:3}(b) \red{A}-\green{C}, \red{D}-\yellow{E}, \red{A}-\red{B}), 
but the cross-patch attention pairs (\red{A}-\green{C}, \red{D}-\yellow{E}, \textit{two pixels from different mini-cubes}) are in far actual distances while intra-patch attention pairs (\red{A}-\red{B}, \textit{two pixels from the same mini-cube}). Therefore, we distinguish the two type of attention pairs and propose the gated relative position biases (\textbf{GRPB}) as shown in Fig.~\ref{fig:5}(b) that uses two learnable real position bias table ($\mathbf{T}_\text{real}$) and pseudo position bias table ($\mathbf{T}_\text{pseudo}$) to replace $\mathbf{T}$. Denote any two pixels in positions $({p},{\hat{p}})$ ($p \in \mathcal{MC}^{k,i,j},\hat p \in \mathcal{MC}^{\hat k,\hat i,\hat j}$), the GRPB between them ($B(p, \hat p)$) can be formulated as



\begin{equation}
\label{eq:grpb}
\mathbf{G}(p, \hat p)=\left\{
\begin{aligned}
1,&~~i=\hat i \wedge j=\hat j \wedge k=\hat k, \\
0,&~~\mathrm{else}
\end{aligned}
\right.
\end{equation}
\begin{equation}
B(p, \hat p) = \mathbf{G}(p, \hat p) \mathbf{T}_\mathrm{real}^{p - \hat p} + (1-\mathbf{G}(p, \hat p)) \mathbf{T}_\mathrm{pseudo}^{p - \hat p}
\label{eq:refer}
\end{equation}
where $p - \hat p$ is the vector difference between the two positions, and used to index the two position bias tables.

\paragraph{Intra-Patch Non-Linear Regression (IP-NLR) Head.} Several recent quality assessment methods \cite{paq2piq,pvq} apply patch-independent regression heads to obtain local quality. Based on the \emph{match constraint} (Eq.~\ref{eq:match}), feature pixels are aligned with mini-cubes, so it is also possible to regress qualities for each mini-cube to obtain local quality maps. Furthermore, as shown in Fig.~\ref{fig:3}(c), the quality-related features in different mini-cubes should be diverse  even in the same video as their original positions are far apart. Therefore, averaging them before regression as commonly practised in video recognition may have the potential risk to lose the sensitivity to the diverse quality information, while regressing them independently can avoid this problem. Based on the two reasons above, we design the Intra-Patch Non-Linear Regression (IP-NLR, Fig.~\ref{fig:5}(c)) to regress the features via a two-layer MLP first and perform pooling on the regressed local quality scores. {Denote final backbone features as $f_\mathrm{final}$, local quality map as $l_{pr}$, the global quality scores (final output of FANet) as $g_{pr}$, linear layers as $\mathrm{L}_1,\mathrm{L}_2$, the IP-NLR can be expressed as follows:}
\begin{align}
    {l}_{pr}^{t,h,w} &= \mathrm{L}_2(\mathrm{GeLU}(\mathrm{L}_1(f_\mathrm{final}^{t,h,w})))   \\ 
    g_{pr} &= \overline{{l}_{pr}} 
\end{align}

\subsection{Adaptive Multi-scale Inference}
\label{section:AMI}

The proposed models can adapt to various computing resources by changing the sampling densities (scales) of \textbf{\textit{fragments}}. However, our conference version \cite{fastvqa} ({FAST-VQA}) still requires training different models for different scales of fragments. This could be inefficient when the input scale needs to be changed frequently, or adaptively. Therefore, with the objective of training at only one scale (\textit{least cost}) and infer at any different scale (\textit{most flexible}), we propose the \textbf{Adaptive Multi-scale Inference (AMI)} for FasterVQA.

To perform AMI, we adaptively modify the backbone structure of FANet with respect to different sizes of inference inputs. Generally, we keep all the linear and pooling layers unchanged as they mainly focus on local textures. For the window-based self-attention layers, we adaptively rescale the attention windows to ensure that the proportion of the window size to the global size is conserved when the input scale changes, which simulates self-attention-based approaches \cite{vit,allyouneed} in dealing with variable-length inputs. Formally, the attention window sizes given new scales of \textbf{\textit{fragments}} are computed as follows:

\begin{equation}
\hat W = \frac{W_{\mathbf{0}} \otimes \hat G}{G_\mathbf{0}}
\end{equation}
where $\hat W$ and $W_{\mathbf{0}}$ are the rescaled and base window sizes, and $\hat G$ and $G_{\mathbf{0}}$ are the actual and preset base number of grids (to meet the \textit{match constraint}, the sizes of mini-cubes are kept the same). For GRPB, we also lookup from the shared $\mathbf{T}^\text{real}$ and $\mathbf{T}^\text{pseudo}$ as defined in Eq.~\ref{eq:refer}, and the gates $\mathbf{G}$ are computed from partitions of actual inputs. Our experiments demonstrate that the proposed FasterVQA with AMI can still infer with high accuracy at a certain scale even without training on \textbf{\textit{fragments}} on the corresponding scale.

\subsection{Objective Functions}
\label{Loss}
Many existing works \cite{rankiqa,qaloss,mdtvsfa} have pointed out that the linearity and monotonicity of quality predictions to ground truth scores are more important objectives than the predictions themselves in quality assessment tasks. Therefore, we define a fusion loss function as the weighted sum of monotonicity loss $\mathcal{L}_{mono}$ and linearity loss $\mathcal{L}_{lin}$ as follows:
\begin{equation}
\mathcal{L}_{mono} = \sum_{i,j} \max((s_{pred}^i - s_{pred}^j)~\textrm{sgn}~(s_{gt}^j - s_{gt}^i), 0)
\end{equation}
\begin{equation}
\mathcal{L}_{lin}  = (1 - \frac{< s_{pred} - \overline{s_{pred}} ,  s_{gt} - \overline{s_{gt}}  >}{\Vert s_{pred} - \overline{s_{pred}} \Vert_2\Vert s_{gt} - \overline{s_{gt}} \Vert_2}) / 2
\end{equation}
\begin{equation}
\mathcal{L}_{fusion}= \mathcal{L}_{lin} + \lambda \mathcal{L}_{mono}
\label{eq:loss}
\end{equation}
where $\mathrm{sgn}(\cdot)$ denotes the sign function, $<>$ denotes the inner product of two vectors, and $s_{pred}$ and $s_{gt}$ are vectors that refer to predictions and ground truth labels in a batch. 



\section{Experiments}

{In the experiment part, we conduct experiments for the proposed concepts and methods in the following aspects:

\vspace{-8pt}
\begin{itemize}
    \item Benchmark comparison with existing approaches (Sec.~\ref{sec:benchmark}), in terms of both accuracy and efficiency.
    \item Detailed evaluation on sampling (Sec.~\ref{sec:sample}), compared to naive sampling approaches and different variants.
    \item Ablation studies on \textit{match constraint}, FANet structure, training and inference strategies, \textit{e.g.} AMI (Sec.~\ref{sec:fanettraining}).
    \item Extra justifications to our methods: irreplaceable role of semantics (Sec.~\ref{sec:rolesemantics}), evaluation on high-resolution cases (Sec.~\ref{sec:hr}) and stability analysis (Sec.~\ref{sec:stab}).
    \item Quantitative studies for local quality maps (Sec.~\ref{sec:vis}).
\end{itemize}}

\subsection{Evaluation Setup}

\subsubsection{Implementation Details} We use the Swin-T~\cite{swin3d} as the backbone of our FANet, which is initialized by pretraining on Kinetics-400 dataset~\cite{k400data}. For FAST-VQA, we implement two sampling densities for \frag~and adjust the window sizes in FANet to the input sizes: \textbf{FAST-VQA} (better accuracy) and \textbf{FAST-VQA-M} (mobile-friendly), as listed in Tab.~\ref{tab:impdetail}. For FasterVQA, as we practice Adaptive Multi-scale Inference (AMI), we unify different sample densities in one single model. Still, we benchmark the performance of FasterVQA on two mobile-friendly scales with reduced size on either spatial (\textbf{FasterVQA-MS}) or temporal (\textbf{FasterVQA-MT}) dimensions together with the base scale (\textbf{FasterVQA}), as listed in Tab.~\ref{tab:impdetail2}. All $S_f$ and $T_f$ are selected to follow the \emph{match constraint}~(Eq.~\ref{eq:match}). The $\lambda$ in Eq.~\ref{eq:loss} is set as $0.3$, with initial learning rate set as $0.001$ for IP-NLR head and $0.0001$ for the Swin-GRPB backbone respectively.

\subsubsection{Evaluation Metrics} We use three metrics, including Pearson Linear Correlation Coefficient ({PLCC}), Spearman Rank-order Correlation Coefficient ({SRCC}), and Kendall Rank-order Correlation Coefficient ({KRCC}), for evaluating the accuracy of quality predictions. PLCC computes the linear correlation between a series of predicted scores and ground truth scores. SRCC will first rank the labels in both series and computes the linear correlation between the two rank series. KRCC computes the rank-pair accuracy, measuring the proportion of correctly predicted relative relations between score pairs.

\begin{table}[]
    \centering
    \setlength\tabcolsep{7.pt}
    \caption{Variants for FAST-VQA with GMS sampling. Both variants require \textbf{4 clips} at inference to cover whole video.}
    \vspace{-12pt}
    \resizebox{\linewidth}{!}{
    \begin{tabular}{l|c|c|c|c|c} \hline
    \makecell[c]{Methods}&\makecell{Number of\\ Frames ($T$)}&\makecell[c]{Size of Mini-patch\\ ($S_f, S_f$)}&\makecell[c]{Number of \\Grids ($G_f$)}&\makecell[c]{Window Size \\ in FANet}&\makecell[c]{FLOPs \\(Infer)}\\\hline
    \textbf{FAST-VQA}&32&(32, 32)&7&(8, 7, 7)&279G\\
    \textbf{FAST-VQA-M}&16&(32, 32)&4&(4, 4, 4)&46G\\
 \hline
    \end{tabular}
    }
    \label{tab:impdetail}
    \vspace{-12pt}
\end{table}

\begin{table}[]
    \centering
    \setlength\tabcolsep{4pt}
    \caption{Inference variants for FasterVQA with St-GMS via AMI.}
    \vspace{-12pt}
    \resizebox{\linewidth}{!}{
    \begin{tabular}{l|c|c|c|c} \hline
    \makecell[c]{Methods}&\makecell[c]{Size of Mini-Cube\\ ($T_f, S_f, S_f$)}&\makecell[c]{Segments and Grids \\ ($G_t, G_s, G_s$)}&\makecell[c]{Rescaled Window Size \\ in FANet ($\hat W$)}&\makecell[c]{FLOPs \\ (Infer)}\\\hline
    \textbf{FasterVQA}&(4, 32, 32)&(8, 7, 7)&(8, 7, 7)&69G\\
    \textbf{FasterVQA-MT}&(4, 32, 32)&(4, 7, 7)&(4, 7, 7)&35G\\
    \textbf{FasterVQA-MS}&(4, 32, 32)&(8, 5, 5)&(8, 5, 5)&36G\\
 \hline
    \end{tabular}
    }
    \label{tab:impdetail2}
    \vspace{-12pt}
\end{table}

\subsubsection{Training \& Benchmark Sets} We use the large-scale LSVQ{$_\text{train}$}\cite{pvq} dataset with 28,056 videos for training FAST-VQA/FasterVQA. For evaluation, we choose 4 testing sets to test the model trained on LSVQ. The first two sets, LSVQ$_\text{test}$ and LSVQ$_\text{1080p}$ are official intra-dataset test subsets for LSVQ, while the LSVQ$_\text{test}$ consists of 7,400 various resolution videos from 240P to 720P, and LSVQ$_\text{1080p}$ consists of 3,600 1080P high resolution videos. We directly evaluate the generalization ability of proposed models on cross-dataset evaluations on KoNViD-1k \cite{kv1k} and LIVE-VQC \cite{vqc}, two widely-recognized in-the-wild VQA benchmark datasets composed of \textit{natural} videos. We also discuss the fine-tuning results on several non-natural VQA datasets, including lab-collected datasets \cite{qualcomm,cvd} and datasets with computer-generated videos \cite{ytugc}, in Sec.~\ref{sec:finetune}. 

\subsection{Benchmark Results}
\label{sec:benchmark}
\subsubsection{Accuracy}

\paragraph{Benchmarking FAST-VQA.} In Tab.~\ref{table:peer}, we compare FAST-VQA with existing classical and deep VQA methods and our baseline, the full-resolution Swin-T with feature regression instead of end-to-end training (denoted as `Full-res Swin-T \textit{feat.}') while it notably outperforms state-of-the-arts with almost ``negligible" cost. FAST-VQA also shows significant improvement to Full-res Swin-T \textit{feat.}, demonstrating that the proposed end-to-end learning via quality-retained sampling is not only much more efficient (with only \textbf{1/42.5} FLOPs required on 1080P videos) but also notably more accurate (with 8.10\% improvement on PLCC metric for LSVQ$_\text{1080p}$) than the existing fixed-feature-based paradigm. 

\begin{figure}[t]
    \centering

        \includegraphics[width=0.93\linewidth]{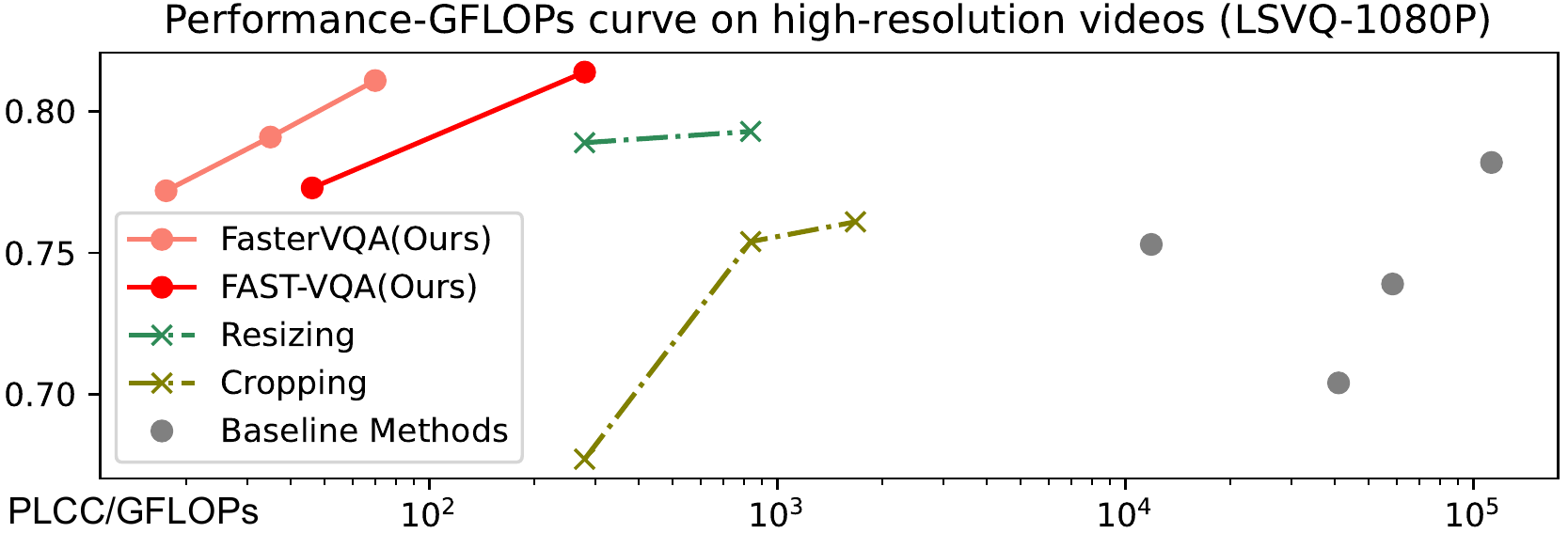}
        \\
        \includegraphics[width=0.93\linewidth]{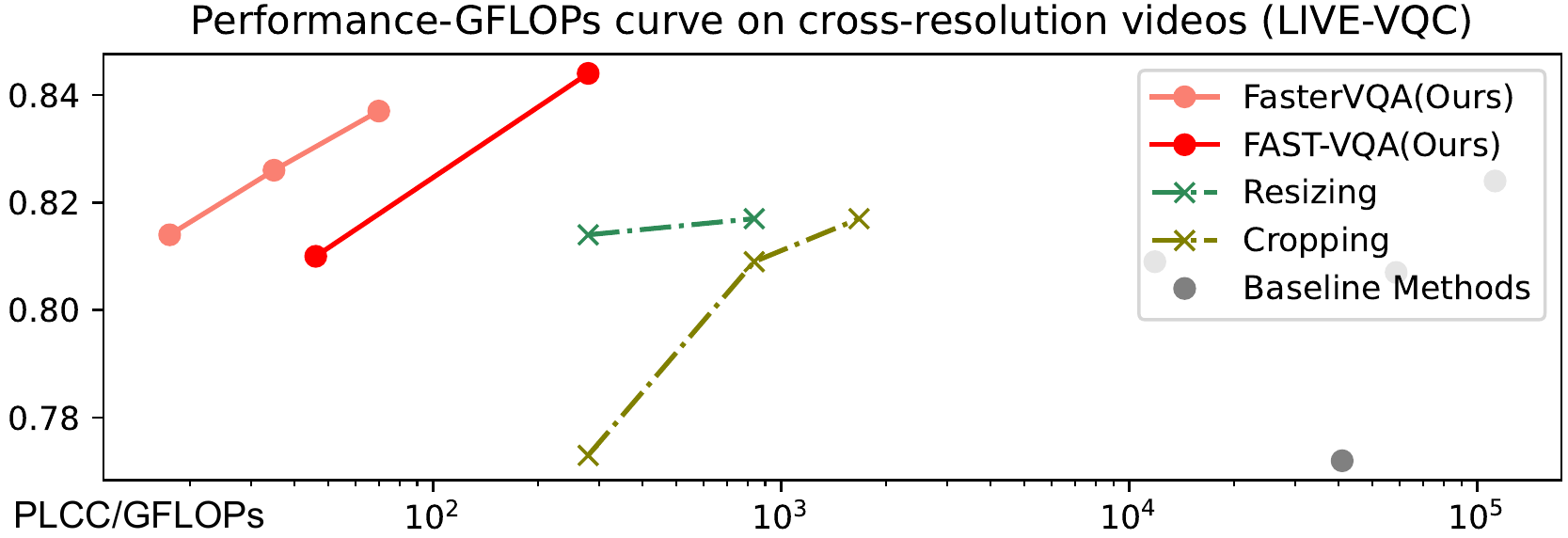}
    \vspace{-10pt}
    \caption{The Performance-FLOPs curve of proposed \green{\textbf{FAST-VQA}} / \green{\textbf{FasterVQA}} and baseline methods. X-Axis: GFLOPs (log scale); Y-Axis: PLCC.}
    \label{fig:a1}
    \vspace{-16pt}
\end{figure}

\label{sec:perf}
\begin{table*}[]
\footnotesize
\caption{Comparison with existing methods (classical and deep) and our baseline (Full-res. Swin-T \textit{feat.}). The 1st/2nd/3rd best scores are colored in \textbf{\red{red}}, \blue{blue} and \textbf{boldface}, respectively. We infer FasterVQA with multiple scales via AMI.} \label{table:peer}
\vspace{-24pt}
\setlength\tabcolsep{7pt}
\renewcommand\arraystretch{1.25}
\footnotesize
\center
\resizebox{\linewidth}{!}{\begin{tabular}{l:l|c|cc|cc|cc|cc}
\hline
\multicolumn{2}{l|}{Type/} & FLOPs on 1080P/8-sec & \multicolumn{4}{c|}{Intra-dataset Test Sets}        &  \multicolumn{4}{c}{Cross-dataset Test Sets}            \\ \hdashline
\multicolumn{2}{l|}{\textbf{Testing Set}/}         & & \multicolumn{2}{c|}{\textbf{LSVQ$_\text{test}$}}   & \multicolumn{2}{c|}{\textbf{LSVQ$_\text{1080p}$}}        &  \multicolumn{2}{c|}{\textbf{KoNViD-1k}}  & \multicolumn{2}{c}{\textbf{LIVE-VQC}}             \\ \hline
Groups~~~ & ~Methods                   & \textit{relative to FAST-VQA} & SRCC & PLCC    & SRCC & PLCC           & SRCC & PLCC                      & SRCC & PLCC                               \\ \hline 
\multirow{3}{0pt}{{{Existing Classical}}} &~BRISQUE\cite{brisque} & NA   & 0.569 &  0.576  & 0.497 & 0.531     & 0.646 &     0.647                   & 0.524 &  0.536 \\
&~TLVQM\cite{tlvqm}  &  NA    & 0.772 &  0.774  & 0.589 & 0.616     & 0.732 &     0.724                   & 0.670 &  0.691 \\
&~VIDEVAL\cite{videval}   &  NA  & 0.794 &  0.783  & 0.545 & 0.554     & 0.751 &     0.741                   & 0.630 &  0.640 \\\hdashline    
\multirow{4}{0pt}{{{Existing \textbf{Fixed} Deep}}} &~VSFA\cite{vsfa} & 147$\times$         & 0.801 &  0.796  & 0.675 & 0.704     & 0.784 &     0.794                   & 0.734 &  0.772 \\

&~PVQ$_\text{wo/ patch}$\cite{pvq}  & 210$\times$ & 0.814 &  0.816  & 0.686 &   0.708         & 0.781 &     0.781                   & 0.747 &  0.776                               \\ 
&~PVQ$_\text{w/ patch}$\cite{pvq} & 210$\times$   & 0.827 &  0.828 & 0.711 &  0.739     & 0.791 &     0.795  & 0.770 &  0.807 \\ 
&~\gray{BVQA-TCSVT-2022}\cite{bvqa2021} & 403$\times$ &  0.852 & 0.854 & \textbf{0.771} & 0.782 & 0.834 & 0.837         &  \blue{0.816} & 0.824 \\ \hdashline
\multicolumn{2}{l|}{Full-res Swin-T\cite{swin3d} \textit{feat.}, $32\times4$ frames} & 42.5$\times$ &  0.835 & 0.833 & 0.739 & 0.753 & 0.825 & 0.828         &  {0.794} & 0.809 \\ \hline
\multirow{3}{0pt}{{{Ours, higher efficiency}}}&{\textbf{FAST-VQA-M}} & {\textbf{0.165$\times$}} &  {{0.852}} & {{0.854}} & {{0.739}} & {{0.773}} & {{0.841}} & {{0.832}} & {0.788} & {{0.810}} \\ 
&{\textbf{FasterVQA-MS} (AMI)} & \textbf{\blue{0.130$\times$}} &  {{0.846}} & {{0.850}}  & {{0.758}} & \textbf{{0.798}} & \textbf{{0.852}} & \textbf{{0.854}}& {{0.791}} & {{0.818}}  \\ 
&{\textbf{FasterVQA-MT} (AMI)} & \textbf{\red{0.125$\times$}} &  \textbf{{0.860}} & \textbf{{0.861}}  & {{0.753}} & {{0.791}} & {{0.846}} & {{0.849}}& {{0.803}} & \textbf{{0.826}}  \\  \hdashline
\multirow{2}{0pt}{{{Ours, Accuracy}}}&{\textbf{FAST-VQA}} & {1$\times$} &  \textbf{\red{0.876}} & \textbf{\red{0.877}}  & \textbf{\red{0.779}} & \textbf{\red{0.814}} & \blue{0.859} & \blue{0.855} & \textbf{\red{0.823}} & \textbf{\red{0.844}}  \\
&{\textbf{FasterVQA}} & {0.25$\times$} &  \blue{{0.873}} & \blue{{0.874}}  & \blue{{0.772}} & \blue{{0.811}} & \textbf{\red{0.863}} & \textbf{\red{0.863}}& \textbf{{0.813}} & \blue{{0.837}} \\  \hline
\end{tabular}}
\vspace{-10pt}

\end{table*}

\begin{table*}[t]
\centering
    \caption{FLOPs and running time (\textit{avg.} of 20 runs) on GPU Server (Tesla V100) and CPU (Apple M1) comparison of FAST-VQA, state-of-the-art methods and our baseline on 8-sec videos different resolutions. We \textbf{boldface} FLOPs $\leq$ 500G, \green{green} FLOPs $\leq$ 100G and running time $\leq$ 1s.} \label{tab:efficiency}
    \vspace{-12pt}
    \setlength\tabcolsep{5pt}
    \renewcommand\arraystretch{1.25}
\resizebox{\linewidth}{!}{
    \begin{tabular}{l|c:c:c|c:c:c|c:c:c}
    \hline
    {} & \multicolumn{3}{c|}{540P} & \multicolumn{3}{c|}{720P} & \multicolumn{3}{c}{1080P}\\
    \cline{2-10}
    {Method} & FLOPs(G) & Time(GPU/s) & Time(CPU/s)  & FLOPs(G) & Time(GPU/s) & Time(CPU/s)  & FLOPs(G) &  Time(GPU/s) & Time(CPU/s) \\
    \hline
     VSFA\cite{vsfa} & 10249\red{$_{36.7\times}$} & 2.603&152.4 & 18184\red{$_{65.2\times}$} & 3.571&233.9 & 40919\red{$_{147\times}$} & 11.14&465.6 \\
    
     PVQ\cite{pvq} & 14646\red{$_{52.5\times}$} & 3.091&149.5 & 22029\red{$_{79.0\times}$} & 4.143&247.8 & 58501\red{$_{210\times}$} & 13.79&538.4 \\
     BVQA-TCSVT-2022\cite{bvqa2021} & 28176\red{$_{101\times}$} & 5.392 & 378.3 & 50184\red{$_{180\times}$} & 10.83 & 592.1 & 112537\red{$_{403\times}$} & 27.64 & 1567 \\ \hdashline
     Full-res Swin-T\cite{swin3d} \textit{feat.} & 3032\red{$_{10.9\times}$} & 3.226&102.0 & 5357\red{$_{19.2\times}$} & 5.049&166.2 & 11852\red{$_{42.5\times}$} & 8.753&234.9 \\ \hline
        \textbf{FAST-VQA} (Ours) & {\textbf{279}{$_{1\times}$}}& \textbfg{0.044}&8.839 & {\textbf{279}{$_{1\times}$}}& \textbfg{0.043}&8.930 & {\textbf{279}{$_{1\times}$}} & \textbfg{0.045}&8.678 \\ 
    \textbf{FasterVQA} (Ours) & {\textbfg{69}\green{$_{0.25\times}$}}& \textbfg{0.023}&2.754 & {\textbfg{69}\green{$_{0.25\times}$}}& \textbfg{0.022}&2.732 & {\textbfg{69}\green{$_{0.25\times}$}} & \textbfg{0.023}&2.697 \\
 \hdashline
    \textbf{FAST-VQA-M} (Ours) & \textbfg{46}\green{$_{0.165\times}$}& \textbfg{0.019}&\textbfg{0.598} & \textbfg{46}\green{$_{0.165\times}$}& \textbfg{0.019}&\textbfg{0.633} & \textbfg{46}\green{$_{0.165\times}$} & \textbfg{0.019}&\textbfg{0.602} \\
    \textbf{FasterVQA-MS} (Ours) & {\textbfg{36}\green{$_{0.130\times}$}}& \textbfg{0.016}&\textbfg{0.594} & {\textbfg{36}\green{$_{0.130\times}$}}& \textbfg{0.018}&\textbfg{0.587}& {\textbfg{36}\green{$_{0.130\times}$}} & \textbfg{0.018}&\textbfg{0.609} \\
    \textbf{FasterVQA-MT} (Ours) & {\textbfg{35}\green{$_{0.125\times}$}}& \textbfg{0.018}&\textbfg{0.647} & {\textbfg{35}\green{$_{0.125\times}$}}& \textbfg{0.020}&\textbfg{0.621}& {\textbfg{35}\green{$_{0.125\times}$}} & \textbfg{0.017}&\textbfg{0.645} \\ \hline
    \end{tabular}
}
    \vspace{-12pt}

\end{table*}

\paragraph{Benchmarking FasterVQA.} We also benchmark the variants of FasterVQA. The base version of FasterVQA achieves performance comparable to FAST-VQA while requiring \textbf{75\% fewer} FLOPs. As FAST-VQA and FasterVQA share the same network structure, the comparison proves the effectiveness of reducing temporal redundancy in VQA in general.
The MS and MT versions of FasterVQA also show notably better performance than FAST-VQA-M, with up to \textbf{24\% fewer} FLOPs. FasterVQA-MT can be more competitive than the recently-published BVQA-TCSVT-2022\cite{bvqa2021} (existing state-of-the-art) in six of eight metrics, while up to \textbf{2,600$\times$} faster. 

\begin{table*}[]
\footnotesize
\caption{The finetune results on LIVE-VQC, KoNViD, CVD2014, LIVE-Qualcomm and YouTube-UGC datasets, compared with existing classical and fixed-backbone deep VQA methods, and ensemble of classical (C) and deep (D) branches.} 
\vspace{-10pt}
\label{table:vqc}
\setlength\tabcolsep{6.6pt}
\renewcommand\arraystretch{1.25}
\footnotesize
\centering
\resizebox{.96\textwidth}{!}{\begin{tabular}{l:l|cc|cc|cc|cc|cc}
\hline
\multicolumn{2}{l|}{\textbf{Finetune Dataset}/}         & \multicolumn{2}{c|}{LIVE-VQC}   & \multicolumn{2}{c|}{KoNViD-1k}        &  \multicolumn{2}{c|}{CVD2014}   &  \multicolumn{2}{c|}{LIVE-Qualcomm}  & \multicolumn{2}{c}{YouTube-UGC}             \\ \hline
\multicolumn{2}{l|}{resolution range in the dataset}  & \multicolumn{2}{c|}{(240P - \textbf{1080P})}   & \multicolumn{2}{c|}{(540P)}        &  \multicolumn{2}{c|}{(480P - 720P)}  & \multicolumn{2}{c}{(\textbf{1080P})} 
 & \multicolumn{2}{c}{(360P - \textbf{2160P(4K)})}             \\ \cline{3-12}
Groups~~~ &~Methods                 & SRCC & PLCC    & SRCC & PLCC           & SRCC & PLCC          &SRCC&PLCC            &SRCC& PLCC                               \\ \hline 
\multirow{3}{0pt}{{{Existing Classical}}} &~TLVQM\cite{tlvqm}        & 0.799 &  0.803  & 0.773 & 0.768     & 0.83 &    0.85               & 0.77 & 0.81 & 0.669 &  0.659 \\

&~VIDEVAL\cite{videval}      & 0.752 &  0.751  & 0.783 & 0.780     & NA &     NA    & NA & NA               & 0.779 &  0.773 \\
&~RAPIQUE\cite{rapique}      & 0.755 &  0.786  & 0.803 & 0.817     & NA &     NA  & NA & NA                 & 0.759 &  0.768 \\\hdashline 
\multirow{5}{0pt}{{{Existing \textbf{Fixed} Deep}}} &~VSFA\cite{vsfa}          & 0.773 &  0.795  & 0.773 & 0.775     & 0.870 &     0.868 & 0.737 & 0.732 & 0.724 &  0.743\\
&~PVQ\cite{pvq}   & {0.827} &  {0.837}  & 0.791 &   0.786       & NA & NA  & NA &     NA                   & NA &  NA\\ 
&~GST-VQA\cite{gstvqa}  & NA &  NA  & 0.814 &   0.825         & 0.831 & 0.844  & 0.801 &  0.825  & NA & NA\\ 
&~CoINVQ\cite{rfugc} & NA &  NA & 0.767 &  0.764  & NA & NA &  NA & NA   & {0.816} &     {0.802}  \\ 
&~BVQA-TCSVT-2022\cite{bvqa2021} & \textbf{0.831} & \textbf{0.842} & 0.834 & 0.836 & 0.872 & 0.869 & \textbf{0.817}& 0.828 & \textbf{0.831} & 0.819 \\\hdashline

\multirow{2}{0pt}{{{Ensemble C+D}}} & CNN+TLVQM\cite{cnntlvqm}        & 0.825 & 0.834 & 0.816 & 0.818 & 0.863 & 0.880  & \textbf{0.810} & 0.833 & NA & NA \\
& CNN+VIDEVAL\cite{videval}        & 0.785 & 0.810 & 0.815 & 0.817 & NA & NA  & NA & NA & 0.808 & \textbf{0.803} \\\hdashline
\multicolumn{2}{l|}{Full-res Swin-T\cite{swin3d} \textit{feat.}} & 0.799 & 0.808 & 0.841 & 0.838 & 0.868 & 0.870 & 0.788 & 0.803 & 0.798 & 0.796 \\ \hline
\multicolumn{2}{l|}{{FAST-VQA-M} (Ours)} & 0.803 & 0.828 & \textbf{0.873} & \textbf{0.872} & \textbf{0.877} & \textbf{0.892} & 0.804 & \textbf{0.838} & 0.768 & 0.765 \\
\multicolumn{2}{l|}{\textit{standard deviation}} &$\pm$.031&$\pm$.030&$\pm$.012&$\pm$.012&$\pm$.035&$\pm$.019&$\pm$.039&$\pm$.026&$\pm$.019&$\pm$.022\\ \hline
\multicolumn{2}{l|}{\textbf{FAST-VQA} (ours)} &  \textbf{\red{0.849}} & \textbf{\red{0.865}} & \textbf{\blue{0.891}} & \textbf{\blue{0.892}} & \textbf{\blue{0.891}} & \textbf{\blue{0.903}} & \textbf{\blue{0.819}} & \textbf{\red{0.851}} & \textbf{\blue{0.855}} & \textbf{\blue{0.852}}  \\
\multicolumn{2}{l|}{\textit{standard deviation}} &$\pm$.024&$\pm$.019&$\pm$.008&$\pm$.008&$\pm$.030&$\pm$.019&$\pm$.036&$\pm$.024&$\pm$.008&$\pm$.011

\\ \hline
\multicolumn{2}{l|}{\textbf{FasterVQA} (ours) \textit{with \red{4X efficiency} than }\textbf{FAST-VQA}} &  \textbf{\blue{0.843}} & \textbf{\blue{0.858}} & \textbf{\red{0.895}} & \textbf{\red{0.898}} & \textbf{\red{0.896}} & \textbf{\red{0.904}} & \textbf{\red{0.826}} & \textbf{\blue{0.844}} & \textbf{\red{0.863}} & \textbf{\red{0.859}} \\
\multicolumn{2}{l|}{\textit{standard deviation}} &$\pm$.032&$\pm$.027&$\pm$.010&$\pm$.010&$\pm$.029&$\pm$.018&$\pm$.038&$\pm$.027&$\pm$.014&$\pm$.017\\ \hline
\end{tabular}}
\vspace{-8pt}
\end{table*}

\subsubsection{Efficiency}
\label{sec:effi}

To benchmark efficiency, we compare the FLOPs and running times on CPU/GPU (average of ten runs per sample) of the proposed methods with existing approaches on different resolutions in Tab.~\ref{tab:efficiency}. We also draw the respective performance-FLOPs curves in Fig.~\ref{fig:a1}. Note that we remove video loading latency for all methods.

\paragraph{Efficiency of base models.}
Even the base models of FAST-VQA and FasterVQA reach unprecedented efficiency. FAST-VQA reduces up to $210\times$ FLOPs and $70\times$ CPU running time than PVQ~\cite{pvq} while obtaining notably better performance, while FasterVQA can reduce up to $840\times$ FLOPs and $284\times$ CPU running time. FasterVQA is also $3.3\times$ faster than FAST-VQA and obviously faster than real-time.

\paragraph{Efficiency of mobile-friendly variants.}
Prior to our submission, the fastest in-the-wild VQA method (including classical methods) on CPU with relatively good accuracy was the RAPIQUE\cite{rapique} model with 17.3s CPU inference time. However, all three of our efficient versions can infer in \textbf{less than one second} on the Apple M1 CPU, which is the processor for several iPad modules. They enable the implementation of more accurate VQA methods on devices with limited computing resources, and we hope the proposed methods can help contribute to green computing on VQA.

\begin{table}
\footnotesize
\caption{Comparsion on ICME2021 UGC-VQA Challenge \cite{icme2021} (Test Set). The results are evaluated by the leaderboard.} \label{table:icme2021}
\vspace{-12pt}
\setlength\tabcolsep{5.5pt}
\renewcommand\arraystretch{1.25}
\footnotesize
\centering
\resizebox{\linewidth}{!}{\begin{tabular}{l:c|cccc}
\hline
Methods & Challenge Rank & SRCC & PLCC & KRCC & RMSE \\ \hline
QA-FTE & 1 & 0.9477 & 0.9831 & 0.8127 & 0.2251 \\
GVSP & 2 & 0.9472 & 0.9809 & 0.8097 & 0.2389 \\
FMISZU & 3 & 0.9471 & 0.9800 & 0.8078 & 0.2441 \\
CENSEO & 4 & 0.9428 & 0.9802 & 0.8020 & 0.2432 \\
\hdashline
\textbf{FAST-VQA (Ours)} & -- & \textbf{\red{0.9552}} & \textbf{\red{0.9878}}& \textbf{\red{0.8266}} & \textbf{\red{{0.1929}}} \\\hline
\end{tabular}}
\vspace{-12pt}
\end{table}

\subsubsection{Fine-tuning on Small Datasets}
\label{sec:finetune}

\paragraph{End-to-end Pre-train\&Fine-tune for VQA.} With \textbf{\textit{fragments}}, we are able to enable the pre-train\&fine-tune scheme for VQA with affordable computational resources, which pre-trains on large VQA datasets to learn quality-related representations and fine-tunes on smaller datasets. This scheme is important as many VQA datasets \cite{vqc,kv1k,ytugc,cvd,qualcomm} in specific scenarios are with much smaller scale than datasets for other video tasks~\cite{sthv2,k400data,activitynet,ava} and it is relatively hard to learn robust quality representations on these small VQA datasets alone. Moreover, the following fine-tuning stage can also be done in an end-to-end manner, which allows the network to learn additional quality-related representations on videos out of the pre-training distributions. 

\paragraph{Results on public datasets.} Practically, we use LSVQ as the large dataset and choose five small datasets representing diverse scenarios, including not only natural video datasets, \textit{i.e.} LIVE-VQC (from real-world mobile photography, 240P-1080P) and KoNViD-1k (from online social media contents, all 540P), but also non-natural datasets: CVD2014 (lab-collected in-capture distortions, 480P-720P), LIVE-Qualcomm (lab-collected videos with specific degradations, all 1080P) and YouTube-UGC (user-generated contents, including computer-generated contents, 360P-2160P\footnote{The current available version of YouTube-UGC is incomplete and only with 1147 videos. The peer comparison is only for reference.}). We divide each dataset into random splits for 10 times and report the average result on the test splits. As Tab.~\ref{table:vqc} shows, with the pre-train\&fine-tune scheme, the proposed FAST-VQA and FasterVQA outperforms the existing state-of-the-arts on all these five scenarios with a very large margin, while obtaining much higher efficiency. Note that YouTube-UGC contains 4K(2160P) videos with 600-frame long but even the FasterVQA still performs well. 

\paragraph{Results on ICME2021 UGC-VQA Challenge.} We also evaluated the fine-tune performance of the proposed FAST-VQA on the ICME2021 UGC-VQA challenge \cite{icme2021}, where the ground truths are hidden and all the methods are fairly evaluated by the challenge server. As shown in Tab.~\ref{table:icme2021}, while the top methods show very similar performance, FAST-VQA is notably better than all of them. As we are not able to pick our model on a hidden-GT database, the result further demonstrates the robustness of FAST-VQA with effective video quality representations.

\subsection{Evaluation on Sampling Approaches}
\label{sec:sample}

\begin{figure}[t]
    \centering
    \includegraphics[width=0.49\columnwidth]{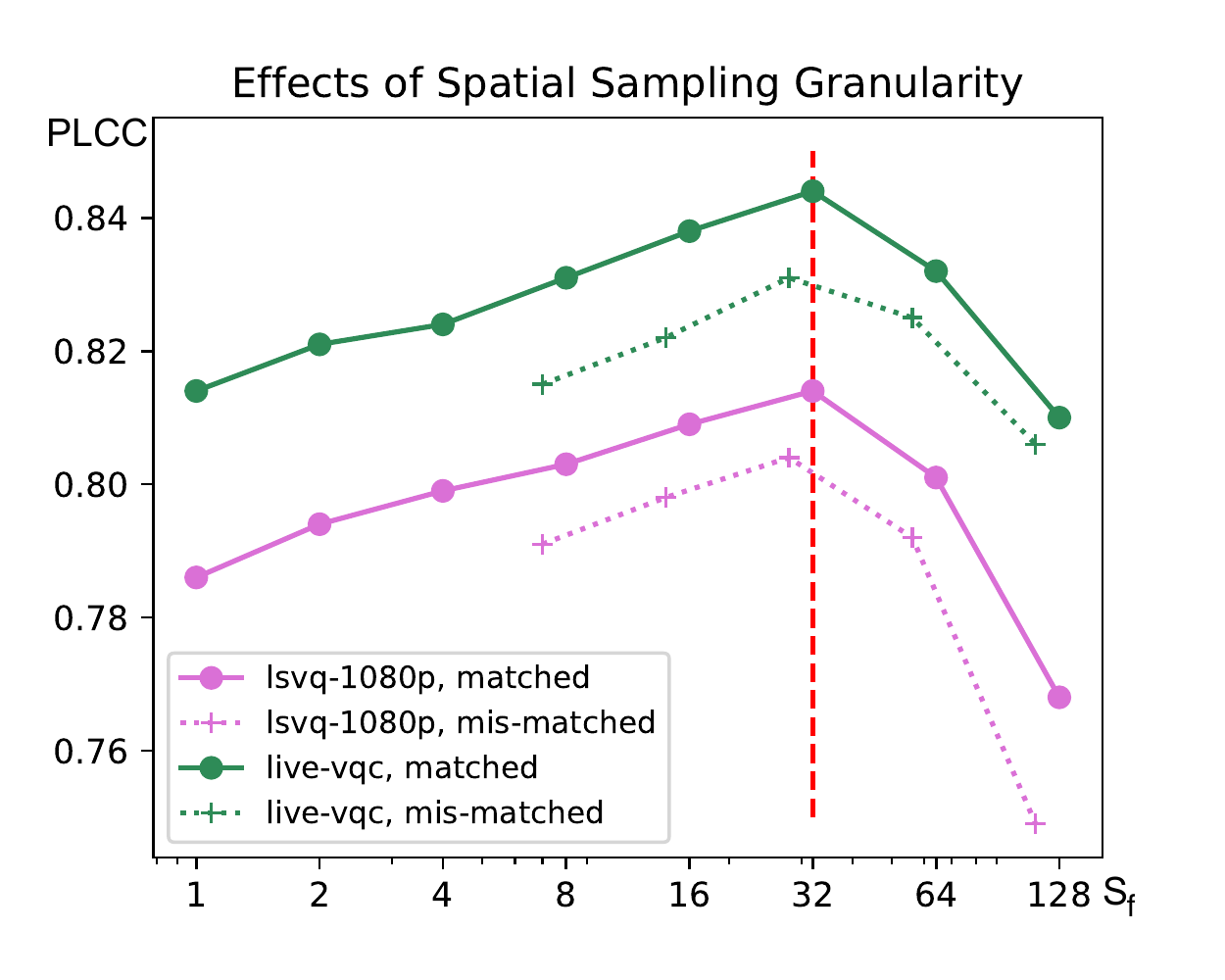} \includegraphics[width=0.49\columnwidth]{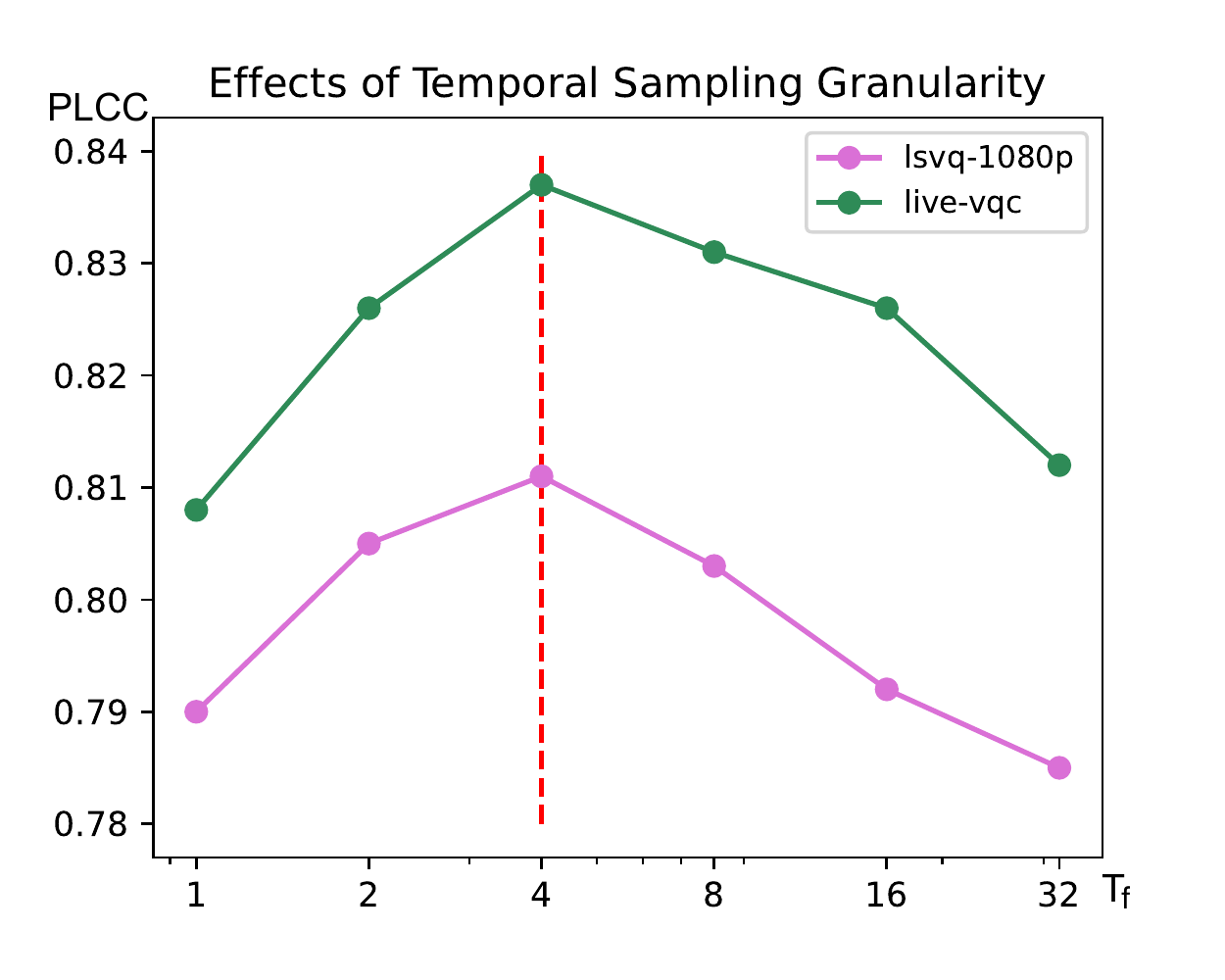}
    \vspace{-9pt}
    \caption{{Discussion on the spatial (in spatial-only GMS) and temporal (in St-GMS) sampling granularity. The dashed lines are for mis-matched combinations, with notably worse performance.}}
    \label{fig:a2}
    \vspace{-15pt}
\end{figure}

We specifically discuss the effects of the proposed sampling paradigm, \textit{quality-sensitive neighbourhood representatives}, and the St-GMS (Sec.~\ref{section:gms}) scheme to get \textbf{\textit{fragments}}. We first show the effectiveness of spatial GMS by comparing it to different spatial sampling variants (Tab.~\ref{tab:resizecrop}), and the effectiveness of unified St-GMS by comparing it to different temporal sampling variants (Tab.~\ref{tab:clipuni}). We also discuss the sampling granularity (Fig.~\ref{fig:a2}) to support the general paradigm of selecting \textit{quality-sensitive neighbourhood representatives}.

\subsubsection{Effects of GMS: in the spatial dimension}

\begin{table*}[]
\footnotesize
\caption{Ablation study for GMS in spatial dimension: comparison with naive approaches and variants.}
\vspace{-13pt}
\setlength\tabcolsep{5pt}
\renewcommand\arraystretch{1.15}
\footnotesize
\centering
\label{tab:resizecrop}
\resizebox{0.87\linewidth}{!}{\begin{tabular}{l|c|cc|cc|cc|cc}
\hline
\textbf{Testing Set}/   &      & \multicolumn{2}{c|}{\textbf{LSVQ$_\text{test}$}}   & \multicolumn{2}{c|}{\textbf{LSVQ$_\text{1080p}$}}        &  \multicolumn{2}{c|}{\textbf{KoNViD-1k}}  & \multicolumn{2}{c}{\textbf{LIVE-VQC}}             \\ \cline{3-10}
Video Resolutions & & \multicolumn{2}{c|}{240p \textit{to} 720p}   & \multicolumn{2}{c|}{1080p}        &  \multicolumn{2}{c|}{540p}  & \multicolumn{2}{c}{240p \textit{to} 1080p}             \\ \cline{3-10}
Methods/Metric   & Relative FLOPs                 & SRCC & PLCC    & SRCC & PLCC           & SRCC & PLCC                      & SRCC & PLCC     \\ \hline     
\multicolumn{10}{l}{{Group 1: Naive Sampling Approaches}} \\ \hdashline
\textit{bilinear resizing} & $1\times$  & 0.857 & 0.859 & 0.752 & 0.786 & 0.841 & 0.840         &  0.772 & 0.814 \\ \hdashline

\textit{random cropping} & $1\times$  & 0.807 & 0.812 & 0.643 & 0.677 & 0.734 & 0.776         &  0.740 & 0.773 \\
- test with 3 crops & $3\times$  & 0.838 & 0.835 & 0.727 & 0.754 & 0.841 & 0.827         &  0.785 & 0.809 \\
- test with 6 crops & $6\times$  & 0.843 & 0.844 & 0.734 & 0.761 & 0.845 & 0.834         &  0.796 & 0.817 \\ \hdashline
\textit{resizing+cropping} with 3 crops, as in \cite{swin3d} & $3\times$ & 0.860 & 0.862 & 0.758 & 0.793 & 0.845 & 0.846         &  0.783 & 0.817 \\
\hline

\multicolumn{10}{l}{{Group 2: Variants of \frag~in the spatial dimension}}    \\ \hdashline                
\textit{random mini-patches} & $1\times$  & 0.857 & 0.861 & 0.754 & 0.790 & 0.844 & 0.845 &  0.792 & 0.818 \\ 
\textit{shuffled mini-patches} & $1\times$ & 0.858 & 0.863 & 0.761 & 0.799 & 0.849 & 0.847 &  0.796 & 0.821 \\ \hdashline
\textit{w/o} temporal alignment & $1\times$  & 0.850 & 0.853 & 0.736 & 0.779 & 0.823 & 0.816         &  0.764 & 0.802 \\
\hline
\textbf{GMS (FAST-VQA, Ours)} & $1\times$  & \textbf{\red{0.876}} & \textbf{\red{0.877}}  & \textbf{\red{0.779}} & \textbf{\red{0.814}} & \textbf{\red{0.859}} & \textbf{\red{0.855}}& \textbf{\red{0.823}} & \textbf{\red{0.844}} \\ \hline
\end{tabular}}
\label{tab:gmstfa}
\footnotesize
\vspace{3pt}
\caption{Ablation study for St-GMS on the temporal dimension: comparison with naive approaches and variants.} 
\vspace{-13pt}
\setlength\tabcolsep{5pt}
\renewcommand\arraystretch{1.15}
\footnotesize
\centering
\label{tab:clipuni}
\resizebox{0.87\linewidth}{!}{\begin{tabular}{l|c|cc|cc|cc|cc}
\hline
\textbf{Testing Set}/    &    & \multicolumn{2}{c|}{\textbf{LSVQ$_\text{test}$}}   & \multicolumn{2}{c|}{\textbf{LSVQ$_\text{1080p}$}}        &  \multicolumn{2}{c|}{\textbf{KoNViD-1k}}  & \multicolumn{2}{c}{\textbf{LIVE-VQC}}             \\ \cline{3-10}
Inter-frame Variations & & \multicolumn{2}{c|}{weak \textit{to} medium}   & \multicolumn{2}{c|}{medium}        &  \multicolumn{2}{c|}{weak}  & \multicolumn{2}{c}{\textbf{strong}}             \\ \cline{3-10}
Temporal Content Changes & & \multicolumn{2}{c|}{medium}   & \multicolumn{2}{c|}{medium}        &  \multicolumn{2}{c|}{\textbf{strong}}  & \multicolumn{2}{c}{weak}             \\ \cline{3-10}
Methods/Metric      & Relative FLOPs           & SRCC & PLCC    & SRCC & PLCC           & SRCC & PLCC                      & SRCC & PLCC     \\ \hline     
\multicolumn{10}{l}{{Group 1: Naive Sampling Approaches}} \\ \hdashline
\textit{sampling a continuous short clip} & $0.25\times$  & 0.853 & 0.856 & 0.750 & 0.785 & 0.833 & 0.834         &  0.782 & 0.812 \\
\hdashline
\textit{uniform sampling (sparse, no continuous frames)} & $0.25\times$  & 0.859 & 0.858 & 0.753 & 0.790 & 0.843 & 0.842         &  0.774 & 0.808 \\ 
\hline
\multicolumn{10}{l}{{Group 2: Variants of \textbf{\textit{fragments}}~in the temporal dimension}}    \\ \hdashline      
\textit{temporally random mini-cubes} & $0.25\times$ & 0.865 & 0.866 & 0.758 & 0.797 & 0.851 & 0.852 &  0.803 & 0.827 \\ 
\textit{temporally shuffled mini-cubes} & $0.25\times$  & 0.864 & 0.866 & 0.756 & 0.793 & 0.853 & 0.854 &  0.807 & 0.828 \\\hline
{\textbf{St-GMS (FasterVQA, ours)} }  & $0.25\times$  & \textbf{\red{0.873}} & \textbf{\red{0.874}}  & \textbf{\red{0.772}} & \textbf{\red{0.811}} & \textbf{\red{0.864}} & \textbf{\red{0.863}}& \textbf{\red{0.813}} & \textbf{\red{0.837}} \\ \hline
\end{tabular}}
\label{tab:stgmst}
\vspace{-11pt}
\end{table*}

\paragraph{Comparing with resizing \& cropping.} In Group 1 of Tab.~\ref{tab:resizecrop}, we compare the proposed fragments with spatial GMS with two common sampling approaches: \textit{bilinear resizing} and \textit{random cropping}. The proposed \textit{fragments} are notably superior to bilinear resizing on \textbf{high-resolution} (LSVQ$_\text{1080p}$) (+4\%) and \textbf{cross-resolution} (LIVE-VQC) scenarios (+4\%). Fragments still lead to non-trivial 2\% improvements over resizing on lower-resolution scenarios where the problems of resizing are not that severe. This proves that keeping local textures is vital for VQA. Fragments also largely outperform single random crops as well as ensembles of multiple crops, suggesting that retaining uniform global quality is also critical to VQA. We additionally compare with Swin-T's original inference samples for video recognition, \textit{resizing+cropping} with three crops, which need 3$\times$ computational cost but still perform notably worse than fragments.

\paragraph{Comparing with spatial variants of fragments.} We also compare with three variants of \frag~in Tab.~\ref{tab:gmstfa}, Group 2. We prove the effectiveness of uniform grid partition by comparing with \textit{random mini-patches} (ignore grids while sampling), and the importance of retaining contextual relations by comparing with \textit{shuffled mini-patches} (sample mini-patches in grids but shuffle them while splicing). The proposed GMS is markedly superior to both variants. Moreover, it shows much better performance than the variant \textit{without} temporal alignment especially on high-resolution videos, indicating that preserving inter-frame temporal variations is necessary for fragments.

\subsubsection{Effects of St-GMS: in the temporal dimension.}

\paragraph{Comparing with uniform \& short-clip sampling.}
In Group 1 of Tab.~\ref{tab:clipuni}, we compare the proposed spatial-temporal fragments with St-GMS in the temporal dimension with two prevalent temporal sampling strategies: \textit{sampling a short clip} and \textit{uniform sampling}. A short clip leads to a notable performance drop on KoNViD-1k\cite{kv1k}, where a non-uniform sample is insufficient to account for the changing content over time. Uniform sampling lacks continuous frames and is especially inaccurate on LIVE-VQC\cite{vqc}, where inter-frame variations are very complicated. The proposed FasterVQA with St-GMS is representative and sensitive to temporal quality and performs better in a variety of situations.

\paragraph{Comparing with temporal variants of fragments.}
Similar to the spatial situation, we also discussed \textit{random} (ignore segments while sampling) and \textit{shuffled mini-cubes}. The results suggest that preserving contextual relations is still important in the temporal dimension and leads to a performance gap of around 1\% across all datasets. However, the gap is notably smaller than in the spatial dimension, indicating that the temporal contextual relations may be less influential on quality than their spatial counterparts.

\subsubsection{Discussion on Sampling Granularity}
\label{sec:granular}

We sample the \textbf{\textit{fragments}} based on the paradigm of quality-sensitive neighbourhood representatives, where we stress two important factors: 1) partitioned neighbourhoods (the more, the better representative); 2) continuous representatives (the larger, the better textural sensitivity). They have to be balanced during practical sampling. We discuss the two important factors by evaluating the spatial and temporal granularity of sampling given a fixed total sample size.

\paragraph{Spatial Granularity: $G_f\&S_f$ in GMS.} We discuss different combinations of number of grids ($G_f$) and size of mini-patches ($S_f$) for GMS, including combinations that follow (solid curves) or not follow (dashed curves) the \emph{match constraint} (Eq.~\ref{eq:match}). We notice that setting $S_f=32$ shows best performance and is better than smaller patches which gradually becomes insensitive to local textures and degenerates into \textit{resizing}), or larger patches which gradually cedes to be representative to global quality and degenerates into \textit{cropping}. (Results of cropping are in Tab.~\ref{tab:resizecrop}). 

\paragraph{Temporal Granularity: $G_t\&T_f$ in St-GMS.}
We also discuss the combinations of number of $G_t$ and $T_f$ for St-GMS given the same total frames. As no temporal pooling is operated in FANet, we only have the matched group, as shown in Fig.~\ref{fig:a2}(b). The $T_f=4$ shows best performance on both datasets which is comparable to dense temporal sampling (FAST-VQA), which follows our observation that a few continuous frames can be sensitive to temporal variations.

\begin{table}
\setlength\tabcolsep{4pt}
\renewcommand\arraystretch{1.15}
\footnotesize
\caption{Ablation study on  backbones: networks that follow the \textit{Match Constraint} are significantly better. All backbones have similar FLOPs ($<$300G).} 
\vspace{-12pt}
\centering
\resizebox{\linewidth}{!}{\begin{tabular}{l|c|c|c|c}
\hline
\textbf{Testing Set}/         & \multicolumn{1}{c|}{\textbf{LSVQ$_\text{test}$}}   & \multicolumn{1}{c|}{\textbf{LSVQ$_\text{1080p}$}}        &  \multicolumn{1}{c|}{\textbf{KoNViD-1k}}  & \multicolumn{1}{c}{\textbf{LIVE-VQC}}             \\ \cline{2-5}
Variants/Metric                   & SRCC/PLCC    & SRCC/PLCC           & SRCC/PLCC                      & SRCC/PLCC     \\ \hline                

\multicolumn{5}{l}{\textbf{``non-matched" backbone} (with overlapping pooling kernels):}    \\ \hdashline 
I3D-ResNet-50 & 0.847/0.846 & 0.717/0.764 & 0.828/0.829 & 0.776/0.808 \\ \hdashline
\multicolumn{5}{l}{\textbf{``matched" backbones} (with non-overlapping pooling kernels):}    \\ \hdashline 
ConvNext-Tiny &  0.869/0.870 & 0.765/0.802 & 0.851/0.852         &  \textbf{\red{0.811}}/\textbf{\red{0.833}} \\
Swin-T (\textit{w/o} GRPB) &  \textbf{\red{0.873}}/\textbf{\red{0.872}} & \textbf{\red{0.769}}/\textbf{\red{0.805}} & \textbf{\red{0.854}}/\textbf{\red{0.853}} & 0.808/0.832 \\ 
\hline
\end{tabular}}
\label{tab:backbone}
\vspace{2pt}
\setlength\tabcolsep{4pt}
\renewcommand\arraystretch{1.15}
\footnotesize
\caption{Ablation study on GRPB and IP-NLR.}
\vspace{-12pt}
\centering
\resizebox{\linewidth}{!}{\begin{tabular}{l|c|c|c|c}
\hline
\textbf{Testing Set}/         & \multicolumn{1}{c|}{\textbf{LSVQ$_\text{test}$}}   & \multicolumn{1}{c|}{\textbf{LSVQ$_\text{1080p}$}}        &  \multicolumn{1}{c|}{\textbf{KoNViD-1k}}  & \multicolumn{1}{c}{\textbf{LIVE-VQC}}             \\ \cline{2-5}
Variants/Metric                   & SRCC/PLCC    & SRCC/PLCC           & SRCC/PLCC                      & SRCC/PLCC     \\ \hline 
\multicolumn{5}{l}{{Variants of \textbf{GRPB}:}}    \\ \hdashline   
\textit{w/o} GRPB (baseline) &  0.873/0.872 & 0.769/0.805 & 0.854/0.853 &  0.808/0.832 \\ 
GRPB on Layers 1\&2 &  0.873/0.875 & 0.772/0.809 & 0.856/0.851 &  0.812/0.838 \\ 
\textit{remove} $\mathbf{T}^\mathrm{pseudo}$ &  0.868/0.869 & 0.763/0.802 & 0.849/0.847 &  0.806/0.831 \\ 
 \hdashline
\multicolumn{5}{l}{{Variants of \textbf{IP-NLR}:}}    \\ \hdashline 
\textit{linear} (baseline) &  0.872/0.873 & 0.768/0.803 & 0.847 /0.849         &  0.810/0.835 \\
\textit{non-linear}, \textit{pool-first}  &  0.873/0.874 & 0.771/0.805 & 0.851/0.850         &  0.813/0.834 \\\hline
\textbf{FANet} (ours) &  \textbf{\red{0.876}}/\textbf{\red{0.877}}  & \textbf{\red{0.779}}/\textbf{\red{0.814}} & \textbf{\red{0.859}}/\textbf{\red{0.855}}& \textbf{\red{0.823}}/\textbf{\red{0.844}}  \\ \hline
\end{tabular}}
\label{tab:netdesign}
\vspace{2pt}
\setlength\tabcolsep{7pt}
\renewcommand\arraystretch{1.15}
\footnotesize
\caption{Ablation study on the Adaptive Multi-scale Inference (AMI) to help inference on different scales.} \centering
\vspace{-12pt}
\resizebox{\linewidth}{!}{\begin{tabular}{l|c|c|c|c}
\hline
\textbf{Testing Set}/         & \multicolumn{1}{c|}{\textbf{LSVQ$_\text{test}$}}   & \multicolumn{1}{c|}{\textbf{LSVQ$_\text{1080p}$}}        &  \multicolumn{1}{c|}{\textbf{KoNViD-1k}}  & \multicolumn{1}{c}{\textbf{LIVE-VQC}}             \\ \cline{2-5}
\multicolumn{5}{l}{Variants of \textbf{FasterVQA-MS:}}    \\ \hdashline 
\textit{without} AMI & 0.838/0.844 & 0.739/0.772 & 0.845/0.842 & 0.782/0.807  \\
\textit{with} AMI &  \bred{{0.846}}/\bred{0.850} &\bred{0.758}/\bred{0.798} & \bred{0.852}/\bred{0.854}         &  \bred{0.791}/\bred{0.818} \\

\hline
\multicolumn{5}{l}{Variants of \textbf{FasterVQA-MT}:}    \\ \hdashline 
\textit{without} AMI &  0.853/0.854 & 0.746/0.782 & 0.841/0.838 & 0.782/0.811 \\
\textit{with} AMI &  \bred{0.861}/\bred{0.860} & \bred{0.753}/\bred{0.791} & \bred{0.846}/\bred{0.849}         &   \bred{0.803}/\bred{0.826} \\

\hline
\end{tabular}}
\label{tab:ablAMI}
\vspace{-11pt}
\end{table}

\subsection{Ablation Studies II on FANet, Training and Inference}
\label{sec:fanettraining}

\subsubsection{Effects of the \emph{Match Constraint}}

\paragraph{Effects of Appropriate Backbones.} In the first part of our ablation studies on FANet, we discuss the effects of different backbone structures by dividing them into two groups: those with non-overlapping pooling layers and can comply with the \emph{match constraint} (Swin-T, inflated ConvNeXt-Tiny) and others (I3D\cite{k400} with ResNet-50 backbone under a modern initialization\cite{resnet50back}). The IP-NLR is included in all variants, while the GRPB is excluded as it is particularly designed for Swin-T. As shown in Tab.~\ref{tab:backbone}, the matched backbones are significantly more effective at processing \textit{\textbf{fragments}} as inputs given similar computational cost, demonstrating our analysis for the \emph{match constraint} (Eq.~\ref{eq:match}).

\paragraph{Effects of Matching Mini-cubes with Pooling.} We further discuss the \emph{match constraint} by comparing the spatial matched (solid lines) \textit{vs} mis-matched mini-cubes (dashed lines) with the same backbone structure. As Fig.~\ref{fig:a2}(a) shows, the non-matched combinations of pooling kernels and mini-cubes show notably worse performance in all situations, again proving the importance of the \emph{match constraint}.

\subsubsection{Effects of GRPB and IP-NLR} In the second part of the ablation studies on FANet, we analyze the effects of two novel modifications in it: the proposed Gated Relative Position Biases (GRPB) and Intra-Patch Non-Linear Regression (IP-NLR) Head as in Tab.~\ref{tab:netdesign}. We compare the IP-NLR with two variants: the linear regression layer and the non-linear regression layers with pooling before regression (\textit{PrePool}). Both modules lead to non-negligible improvements especially on high-resolution (LSVQ$_\text{1080p}$) or cross-resolution (LIVE-VQC) scenarios. As  the discontinuity between mini-patches is more obvious in high-resolution videos, this result suggests that the corrected position biases and regression head are helpful on solving the problems caused by such discontinuity.

\subsubsection{Effects of Adaptive Multi-scale Inference (AMI)}

In the third part, we evaluate the importance of Adapive Multi-scale Inference (AMI) to allow inference of FasterVQA on different scales with only training on one base scale. In Tab.~\ref{tab:ablAMI}, we evaluate the inference accuracy on MT and MS scales with or without AMI. The results have demonstrated the effectiveness of AMI, which allows robust inference on multiple scales for different test sets.


\subsubsection{Effects of End-to-end Pre-train\&Fine-tune Scheme}

We discuss the effects of pre-train\&fine-tune scheme (Sec.~\ref{sec:finetune}) in Tab.~\ref{table:vqr} in comparison with direct training on these small datasets (\textit{w/o} end-to-end pre-train) and only linear regression on pre-trained features (\textit{w/o} end-to-end finetune). The large-scale pre-training contributes to the performance by up to 11\%, and are especially effective on cross-resolution scenarios, \textit{e.g.} LIVE-VQC and YouTube-UGC. The end-to-end fine-tune also lead to up to 8\% improvements, especially on non-natural videos (CVD2014, LIVE-Qualcomm, YouTube-UGC) which may contain specific quality-related issues. Both stages are undoubtedly effective and made affordable via the proposed \textit{fragments}.

\begin{table*}
\footnotesize
\caption{Effects of end-to-end pre-training and fine-tuning processes on downstream small VQA datasets.} \label{table:vqr}
\vspace{-12pt}
\setlength\tabcolsep{8pt}
\renewcommand\arraystretch{1.1}
\footnotesize
\centering
\resizebox{0.85\linewidth}{!}{\begin{tabular}{l:l|cc|cc|cc|cc|cc}
\hline
\multicolumn{2}{l|}{\textbf{Finetune Dataset}/}         & \multicolumn{2}{c|}{LIVE-VQC}   & \multicolumn{2}{c|}{KoNViD-1k}        &  \multicolumn{2}{c|}{CVD2014}   &  \multicolumn{2}{c|}{LIVE-Qualcomm}  & \multicolumn{2}{c}{YouTube-UGC}             \\ \hline
\multicolumn{2}{l|}{\textbf{Metric}}                 & SRCC & PLCC    & SRCC & PLCC           & SRCC & PLCC               &SRCC& PLCC   &SRCC&PLCC                           \\ \hline
\multicolumn{2}{l|}{ \textit{w/o} end-to-end pre-train } & 0.765 & 0.782 & 0.842 & 0.844 & 0.871 & 0.888 & 0.756 & 0.778 & 0.794 & 0.784 \\ \hdashline
\multicolumn{2}{l|}{ \textit{w/o} end-to-end fine-tune } & 0.818 & 0.838 & 0.869 & 0.868 & 0.822 & 0.840 & 0.740 & 0.787 & 0.814 & 0.811 \\ \hdashline
\multicolumn{2}{l|}{\textbf{FAST-VQA} (ours)} &  \textbf{\red{0.849}} & \textbf{\red{0.865}} & \textbf{\red{0.891}} & \textbf{\red{0.892}} & \textbf{\red{0.891}} & \textbf{\red{0.903}} & \textbf{\red{0.819}} & \textbf{\red{0.851}} & \textbf{\red{0.855}} & \textbf{\red{0.852}}  \\ \hline
\end{tabular}}
\vspace{-17pt}
\end{table*}

\subsection{Role of Semantics in FAST-VQA/FasterVQA}

\label{sec:rolesemantics}

\paragraph{Can fragments preserve semantics?} In our discussions in Sec.~\ref{section:stgms}, one question remains unclear: can the \textbf{\textit{fragments}} retain aware to semantic video contents that can still be recognized by deep neural networks? This can hardly be answered as for a 10-sec-long 720P video, \textbf{\textit{fragments}} sampled by St-GMS contain only 0.58\% original information. Thus, we measure the ability by experiments: we use fragments as classification inputs for videos in Kinetics-400\cite{k400data} action recognition dataset, and the results prove that simply fine-tuning the Swin-T backbone with fragments can reach \textbf{68.6\%} top-1 accuracy (87.4 \% relative to original Swin-T which needs 12 samples and requires 12$\times$ FLOPs) and \textbf{88.7\%} top-5 accuracy (94.8\% relative to original), which has been on par with several deep VQA approaches under similar computational cost. The absolute accuracy also suggests that the \textbf{\textit{fragments}} still contain rough scene-level semantics and can be recognized by the backbone in FANet.

\begin{table}
\setlength\tabcolsep{6pt}
\renewcommand\arraystretch{1.15}
\footnotesize
\caption{Effects of semantic pre-training on Kinetics-400.} 
\vspace{-13pt}
\centering
\resizebox{\linewidth}{!}{\begin{tabular}{l|c|c|c|c}
\hline
\textbf{Testing Set}/         & \multicolumn{1}{c|}{\textbf{LSVQ$_\text{test}$}}   & \multicolumn{1}{c|}{\textbf{LSVQ$_\text{1080p}$}}        &  \multicolumn{1}{c|}{\textbf{KoNViD-1k}}  & \multicolumn{1}{c}{\textbf{LIVE-VQC}}             \\ \cline{2-5}
Variants/Metric                   & SRCC/PLCC    & SRCC/PLCC           & SRCC/PLCC                      & SRCC/PLCC     \\ \hline 
\multicolumn{5}{l}{\textbf{Existing Classical Methods:}}    \\ \hdashline
VIDEVAL\cite{videval} 	& 0.794/0.783 & 0.545/0.554	& 0.751/0.741 &	0.630/0.640 \\ 
TLVQM\cite{tlvqm} &	0.772/0.774  &  0.589/0.616 &	0.734/0.724 &	0.670/0.690 \\ \hdashline
\multicolumn{5}{l}{\textbf{FAST-VQA:}}    \\ \hdashline   
\textit{w/o} semantics & 0.788/0.791  &  0.662/0.707 & 0.802/0.793 &  0.737/0.766 \\ 
\textit{w/} semantics &  0.876/0.877 & 0.779/0.814 & 0.859/0.855 &  0.823/0.844 \\ 
 \hdashline
\multicolumn{5}{l}{{\textbf{FasterVQA:}}}    \\ \hdashline 
\textit{w/o} semantics &	0.763/0.760	 & 0.634/0.685 & 0.770/0.778 & 0.720/0.739 \\ 
\textit{w/} semantics &  0.873/0.874 & 0.772/0.811 & 0.863/0.864 &  0.813/0.837 \\ \hline
\end{tabular}}
\label{tab:sempretrain}
\end{table}

\paragraph{Effects of Semantic Pre-training.} We further discuss the significance of semantic pre-training by training FAST-VQA/FasterVQA models from scratch (\textbf{w/o} semantics) as their semantic-blind variants, and the proposed models are regarded as semantic-aware (\textit{w/} semantics) variants based on discussions above. As shown in Tab.~\ref{tab:sempretrain}, semantic pre-training has significantly contributed to the performance on FAST-VQA (\textit{avg.} 8\%) and FasterVQA (\textit{avg.} 10\%), especially FasterVQA. We also observed that the intra-dataset performance of the state-of-the-art classical VQA approaches is comparable to that of our variants without semantic pre-training. The results indicate the significant influence of semantics in VQA and suggest that there might exist an accuracy limit of all semantic-blind VQA methods. This further proves that semantic-aware deep VQA methods are irreplaceable, while FAST-VQA and FasterVQA fill in the blanks on improving their practical efficiency.

\subsection{Evaluation on High-resolution Videos}
\label{sec:hr}

As the base version of FAST-VQA only samples 5.44\% and 2.42\% spatial information from 720P and 1080P videos, respectively, it is worthwhile to evaluate its performance on high-resolution videos. We use two existing databases with 1080P videos: for cross-resolution LIVE-VQC, we split the videos according to their resolutions and test the performance of different variants; for LSVQ$_\text{1080p}$, we create variants by downsampling its 1080P videos before sampling \textbf{\textit{fragments}} and compare between them.
\subsubsection{Performance on Split Resolutions}
    We divide the cross-resolution VQA benchmark set LIVE-VQC into three resolution groups: (A) 1080P (110 videos); (B) 720P (316 videos); and (C) $\leq$540P (159 videos) to evaluate the performance of FAST-VQA on different resolutions in comparison to other variants. As shown in Tab.~\ref{tab:resolution}, the proposed FAST-VQA achieves good performance on all resolution groups ($\geq$0.80 SRCC\&PLCC), with the most superior improvement over other variants on Group (A) with 1080P high-resolution videos, proving that FAST-VQA is robust and reliable on videos with different resolutions.

\begin{table}
\center
\setlength\tabcolsep{3pt}
\renewcommand\arraystretch{1.15}
\footnotesize
\vspace{-10pt}
\caption{Performance on split resolutions of LIVE-VQC.} \vspace{-20pt}
\resizebox{\linewidth}{!}{\begin{tabular}{l|c|c|c}
\hline
\textbf{Resolution}     & \multicolumn{1}{c|}{(A): 1080P}   & \multicolumn{1}{c|}{(B): 720P}        &  \multicolumn{1}{|c}{(C): $\leq$540P}    \\ \cline{2-4}
Variants                    & SRCC/PLCC/KRCC    & SRCC/PLCC/KRCC         & SRCC/PLCC/KRCC     \\ \hline    
\textit{Full-res} Swin \textit{features} & 0.771/0.774/0.584 & 0.796/0.811/0.602 & 0.810/0.853/0.625 \\ \hdashline
\textit{bilinear resizing} & 0.758/0.773/0.573 & 0.790/0.822/0.599 & 0.835/0.878/0.650 \\ \hdashline
\textit{random cropping} & 0.765/0.768/0.565 & 0.774/0.787/0.581 & 0.730/0.809/0.535 \\  \hdashline

\textit{w/o} GRPB & 0.796/0.785/0.598 & 0.802/0.820/0.608 & 0.834/0.883/0.649 \\ \hline
\textbf{FAST-VQA} (Ours) & \bred{0.807}/\bred{0.806}/\bred{0.610} & \bred{0.803}/\bred{0.825}/\bred{0.610} & \bred{0.840}/\bred{0.885}/\bred{0.654} \\ \hline

\end{tabular}}
\label{tab:resolution}
\vspace{-9pt}
\end{table}

\begin{figure}[t]
    \centering
    \includegraphics[width=0.99\linewidth]{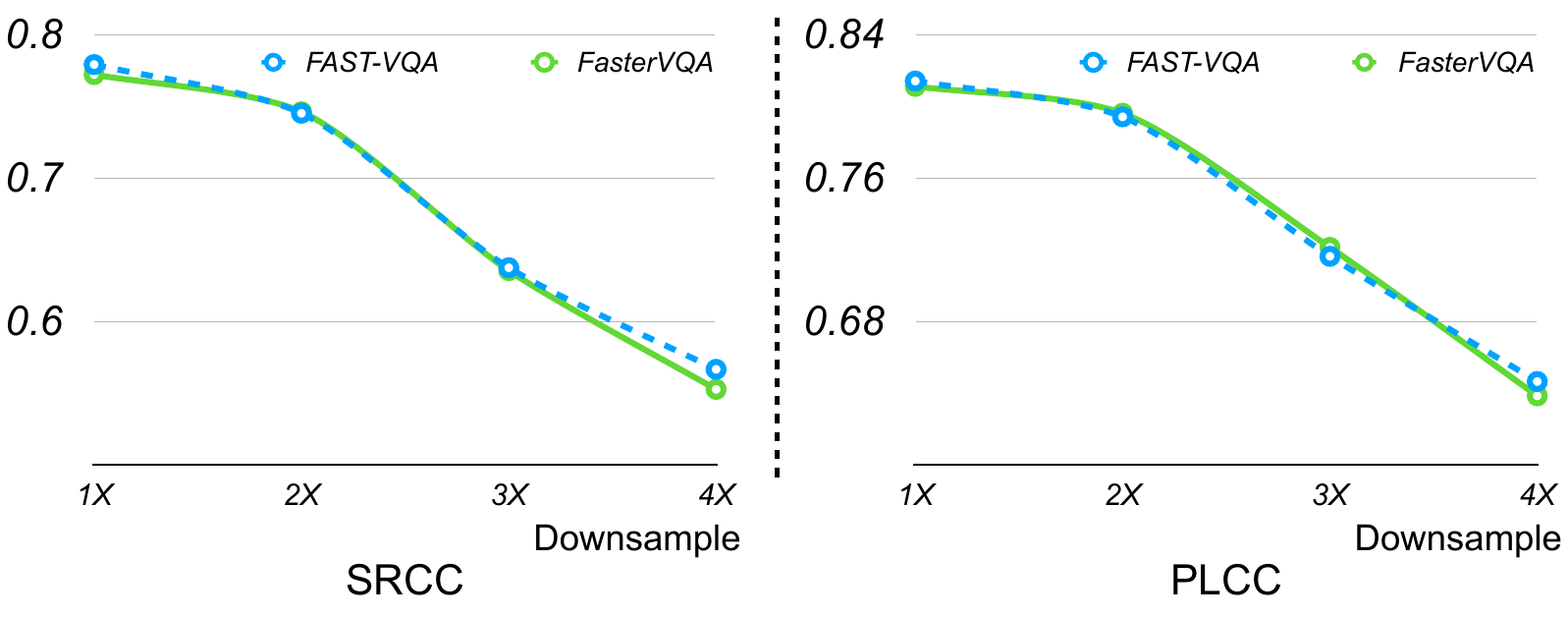}
    \vspace{-12pt}
    \caption{Impacts of downsampling 1080P videos in LSVQ$_\text{1080P}$.}
    \label{fig:cres}
    \vspace{-5pt}
\end{figure}

\begin{figure*}[]
    \centering
    \includegraphics[width=0.96\linewidth]{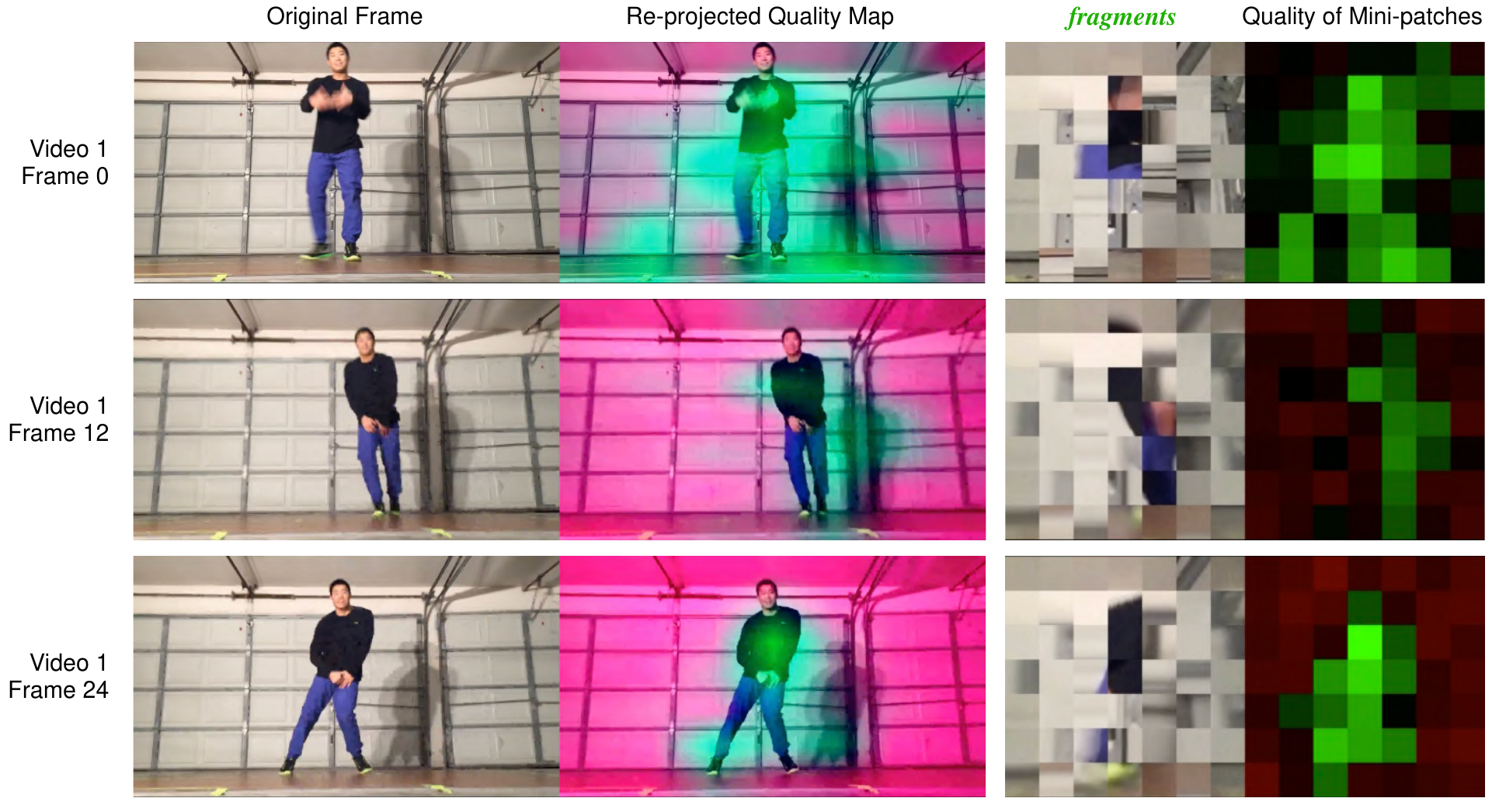}
    \vspace{-14pt}
    \caption{Spatial-temporal patch-wise local quality maps, where \textbf{\red{red}} areas refer to low predicted quality and \textbf{\green{green}} areas refer to high predicted quality. This sample video is a 1080P video from LIVE-VQC~\cite{vqc} dataset. Zoom in for clearer view.}
    \label{fig:6}
    \vspace{-12pt}
\end{figure*}

\subsubsection{Impacts of Video Downsampling}

To demonstrate that keeping the raw-resolution textures is crucial in sampling \textbf{\textit{fragments}}, we evaluate the proposed FAST-VQA/FasterVQA with multiple downsampled variants of LSVQ$_\text{1080p}$ dataset. We resize these 1080P high-resolution videos into 540P(2X$\downarrow$), 360P(3X$\downarrow$), 270P(4X$\downarrow$) and sample \textit{{fragments}} from the resized videos. As shown in Fig.~\ref{fig:cres}, although downsampling before sampling can preserve more information from these videos, the overall effect still significantly degrades the final accuracy, proving that keeping the original resolution is crucial to quality sensitivity. As the model is only trained on videos $\leq$720P, the result further reveals the general importance of textures on different resolutions of videos.

\subsection{Stability and Reliability Analysis}
\label{sec:stab}

Due to the randomness of fragment sampling, the proposed FAST-VQA may produce varying predictions for the same video. Therefore, we measure the stability and reliability of single random sampling in FAST-VQA using two metrics: 1) the assessment stability of multiple single samplings on the same video; 2) the relative accuracy of single sampling compared with multiple sample ensemble. As shown in Tab.~\ref{tab:stability}, the normalized \textit{std. dev.} of different sampling on the same video is only around 0.01, indicating that the sampled fragments are enough for making highly stable predictions. Compared with a six-sample ensemble, sampling only once can be 99.40\% as accurate even on the pure high-resolution test set (LSVQ$_\text{1080P}$). They prove that a single sample of \frag~is sufficiently stable and reliable for quality assessment even though only a small proportion of information is kept during sampling.

\begin{table}
\center
\setlength\tabcolsep{6pt}
\renewcommand\arraystretch{1.15}
\footnotesize
\vspace{-8pt}
\caption{Stability and reliability of single sampling of \frag~.}
\vspace{-20pt}
\resizebox{\linewidth}{!}{\begin{tabular}{l|c|c|c|c}
\hline
\textbf{Testing Set}/         & \multicolumn{1}{|c|}{\textbf{LSVQ$_\text{test}$}}   & \multicolumn{1}{|c|}{\textbf{LSVQ$_\text{1080p}$}}        &  \multicolumn{1}{|c|}{\textbf{KoNViD-1k}}  & \multicolumn{1}{|c}{\textbf{LIVE-VQC}}             \\ \cline{2-5}
Score Range & 0-100 & 0-100 & 1-5 & 0-100 \\ \hline
\textit{std. dev.} of Single Samplings  & 0.65 & 0.79 & 0.046 & 1.07 \\ 
Normalized \textit{std. dev.} & 0.0065 & 0.0079 & 0.0115 & 0.0107 \\ \hline
\textit{Avg.} KRCC on Single Sampling  & 0.6918 & 0.5862 & 0.6693 &  0.6296 \\
KRCC on 6-sample ensemble & 0.6947 & 0.5897 & 0.6730 & 0.6326  \\ \hdashline
Relative Accuracy & 99.59\% & 99.40\% & 99.45\% & 99.52\%  \\ \hline
\end{tabular}}
\label{tab:stability}
\vspace{-8pt}
\end{table}

\subsection{Visualizations of Local Quality Maps}
\label{sec:vis}

The proposed IP-NLR head with patch-wise independent quality regression not only improves the performance of the proposed method but also enables the generation of spatial-temporal local quality maps as \cite{pvq} does. These quality maps allow us to qualitatively evaluate what can be learned during the end-to-end training for FAST-VQA. We show the patch-wise local quality maps and the re-projected frame quality maps for a 1080P video (from LIVE-VQC~\cite{vqc} dataset) in Fig.~\ref{fig:6}. As the patch-wise quality maps and re-projected quality maps in Fig.~\ref{fig:6} (column 2\&4) shows, FAST-VQA is sensitive to textural quality information and distinguishes between clear (Frame 0) and blurry textures (Frame 12/24). It demonstrates that FAST-VQA with \frag~(column 3) as input is sensitive to local texture quality. Furthermore, the qualities of the action-related areas are notably different from those of the background areas, showing that FAST-VQA effectively learns the global contextual relations. It is aware of and influenced by semantic information in the video, thereby demonstrating our aforementioned claims. More visualizations of local quality maps are presented in our GitHub page, together with codes and models.

\section{Conclusions}

In this paper, we have discussed sampling for video quality assessment (VQA) in order to tackle the difficulties as a result of high computing and memory requirements when evaluating high-resolution videos. We propose the principle of quality-sensitive neighbourhood representatives and conduct extensive experiments to demonstrate that the proposed samples, \frag, are effective samples for VQA that retain quality information in videos better than naive sampling approaches. Based on \frag, the proposed end-to-end FAST-VQA and FasterVQA refreshed state-of-the-arts on all in-the-wild VQA benchmarks with up to 1612$\times$ efficiency than the existing state-of-the-art. The proposed methods can bring deep VQA methods into practical use regardless of video resolution or length. In our future work, we would like to further improve specific network structures with insights from the \textit{match constraint} and design more effective sampling approaches based on the principle of quality-sensitive neighbourhood representatives. 

\vspace{-10pt}
\bibliographystyle{IEEEtran}
\bibliography{egbib}

\end{document}